\documentclass[letterpaper]{article} 
\usepackage{aaai2027}  
\usepackage[hyphens]{url}  
\usepackage{graphicx} 
\usepackage{natbib}  
\usepackage{caption} 
\usepackage{algorithm}
\usepackage{algorithmic}
\usepackage{amsmath}
\usepackage{amssymb}
\usepackage{mathrsfs}
\usepackage{enumitem}
\usepackage{booktabs}
\usepackage{multirow}
\usepackage{graphicx}
\usepackage[table]{xcolor}   
 \usepackage{bm}
\definecolor{oursrow}{RGB}{232,240,254}  
\definecolor{oursrow}{RGB}{232,240,254}
\definecolor{up}{RGB}{0,120,60}      
\definecolor{down}{RGB}{180,30,30}   
\usepackage{newfloat}
\usepackage{listings}
\DeclareCaptionStyle{ruled}{labelfont=normalfont,labelsep=colon,strut=off} 
\floatstyle{ruled}
\newfloat{listing}{tb}{lst}{}
\floatname{listing}{Listing}
\usepackage{booktabs}
\usepackage{multirow}
\usepackage{booktabs}
\usepackage{xspace}
\definecolor{headerbg}{HTML}{15324B}   
\definecolor{grouprbg}{HTML}{2E5E8C}   
\definecolor{groupmbg}{HTML}{3A7150}   
\definecolor{ourshl}{HTML}{FFF1CC}     
\definecolor{bestcol}{HTML}{0B4F9E}    
\definecolor{ablbest}{HTML}{7A3B12}    

\makeatletter
\g@addto@macro{\normalsize}{%
\setlength{\abovedisplayskip}{1.4pt plus1pt}%
\setlength{\abovedisplayshortskip}{1.4pt plus1pt}%
\setlength{\belowdisplayskip}{1.4pt plus1pt}%
\setlength{\belowdisplayshortskip}{1.4pt plus1pt}}
\makeatother
\newcommand{\best}[1]{\textbf{\textcolor{bestcol}{#1}}}      
\newcommand{\abest}[1]{\textbf{\textcolor{ablbest}{#1}}}     
\newcommand{\snd}[1]{\underline{#1}}                          
\newcommand{\ourslong}{\textbf{\underline C}itation-guided \textbf{\underline R}esearch \textbf{\underline A}gent for \textbf{\underline S}cholarly \textbf{\underline E}xploration\xspace}
\definecolor{oursaccent}{RGB}{45,90,160}
\newcommand{\ours}{\textcolor{oursaccent}{\textsc{Crase}}}
\newcommand{\gpt}{DeepResearch-GPT\xspace}
\newcommand{\claude}{DeepResearch-Claude\xspace}
\newcommand{\spector}{Spector2-Deepwalk\xspace}

\usepackage[most]{tcolorbox}
\usepackage{xcolor}
\definecolor{cataccent}{RGB}{45,90,160}
\newcommand{\cat}[1]{\textcolor{cataccent}{\texttt{#1}}}
\newtcolorbox{promptbox}[1]{
    enhanced,
    breakable,
    colback=black!2,
    colframe=black!35,
    boxrule=0.6pt,
    arc=2pt,
    left=6pt,
    right=6pt,
    top=6pt,
    bottom=5pt,
    title=\textbf{#1},
    coltitle=black,
    colbacktitle=black!7,
    fonttitle=\small,
    attach boxed title to top left={
        xshift=6pt,
        yshift=-2.5mm
    },
    boxed title style={
        boxrule=0pt,
        arc=1pt,
        left=3pt,
        right=3pt,
        top=1pt,
        bottom=1pt
    }
}

\usepackage{xcolor}
\usepackage{colortbl}
\usepackage{soul}       
\usepackage{pifont}     
\definecolor{keptteal}{RGB}{0,109,119}
\definecolor{prunedgray}{RGB}{145,145,145}
\definecolor{grouphead}{RGB}{28,60,110}
\definecolor{groupbg}{RGB}{236,240,247}
\newcommand{\kept}[1]{\textcolor{keptteal}{#1}}
\newcommand{\pruned}[1]{\textcolor{prunedgray}{\st{#1}}}
\newcommand{\keptmark}{\textcolor{keptteal}{\ding{51}}}
\newcommand{\prunedmark}{\textcolor{prunedgray}{\ding{55}}}
\newcommand{\subq}[2]{\rowcolor{groupbg}\multicolumn{2}{l}{%
\textcolor{grouphead}{\textbf{#1}\;\;\textit{``#2''}}} \\}

\usepackage[most]{tcolorbox}
\usepackage{xcolor}

\definecolor{tracebg}{RGB}{248,249,250}
\definecolor{tracekey}{RGB}{136,57,239}   
\definecolor{traceval}{RGB}{64,120,192}   
\definecolor{tracenote}{RGB}{192,80,20}   
\definecolor{traceframe}{RGB}{60,60,60}

\newtcolorbox{tracebox}[1]{
  enhanced, breakable,
  colback=tracebg, colframe=traceframe, boxrule=0.6pt, arc=1.5mm,
  left=2mm, right=2mm, top=4mm, bottom=1.5mm,
  title={\small\ttfamily\bfseries #1}, fonttitle=\color{white},
  attach boxed title to top left={yshift=-2mm, xshift=3mm},
  boxed title style={colback=traceframe, arc=1mm}
}
\newcommand{\tkey}[1]{\textcolor{tracekey}{\ttfamily\bfseries #1}}
\newcommand{\tval}[1]{\textcolor{traceval}{\ttfamily #1}}
\newcommand{\tnote}[1]{{\textcolor{tracenote}{\itshape\footnotesize \ $\triangleleft$ #1}}}

\usepackage[most]{tcolorbox}

\definecolor{rqaccent}{RGB}{45,90,160}   

\newtcolorbox{researchquestion}{
  enhanced,
  breakable,
  colback=rqaccent!4,      
  colframe=rqaccent,       
  boxrule=0pt,             
  leftrule=2.5pt,          
  arc=1.5pt,
  left=6pt, right=6pt, top=4pt, bottom=4pt,
  fonttitle=\bfseries\footnotesize,
  coltitle=rqaccent,
  attach title to upper={\par\medskip},
}

\usepackage[most]{tcolorbox}
\usepackage{xcolor}
\definecolor{cataccent}{RGB}{45,90,160}

\usepackage{xcolor}
\usepackage{colortbl}
\usepackage{soul}       
\usepackage{pifont}     
\definecolor{keptteal}{RGB}{0,109,119}
\definecolor{prunedgray}{RGB}{145,145,145}
\definecolor{grouphead}{RGB}{28,60,110}
\definecolor{groupbg}{RGB}{236,240,247}

\usepackage[most]{tcolorbox}
\usepackage{xcolor}

\definecolor{tracebg}{RGB}{248,249,250}
\definecolor{tracekey}{RGB}{136,57,239}   
\definecolor{traceval}{RGB}{64,120,192}   
\definecolor{tracenote}{RGB}{192,80,20}   
\definecolor{traceframe}{RGB}{60,60,60}

\usepackage[most]{tcolorbox}

\definecolor{rqaccent}{RGB}{45,90,160}   

\usepackage[most]{tcolorbox} \usepackage{xcolor} \definecolor{faqblue}{RGB}{235,244,252} \definecolor{faqblueframe}{RGB}{62,119,170} \definecolor{faqgreen}{RGB}{237,248,242} \definecolor{faqgreenframe}{RGB}{60,140,100} \definecolor{faqorange}{RGB}{253,244,232} \definecolor{faqorangeframe}{RGB}{190,125,55} \newtcolorbox{faqbluebox}{ colback=faqblue, colframe=faqblueframe, boxrule=0.6pt, arc=2pt, left=5pt,right=5pt,top=4pt,bottom=4pt, before skip=5pt,after skip=5pt } \newtcolorbox{faqgreenbox}{ colback=faqgreen, colframe=faqgreenframe, boxrule=0.6pt, arc=2pt, left=5pt,right=5pt,top=4pt,bottom=4pt, before skip=5pt,after skip=5pt } \newtcolorbox{faqorangebox}{ colback=faqorange, colframe=faqorangeframe, boxrule=0.6pt, arc=2pt, left=5pt,right=5pt,top=4pt,bottom=4pt, before skip=5pt,after skip=5pt }

\usepackage{changepage}   
\usepackage{graphicx}     
\usepackage{url}          

\newcommand{\opensource}{%
    \begin{adjustwidth}{3pt}{3pt}
        \begin{center}
            \vspace{0.1cm}
            
            \raisebox{-0.15em}{\includegraphics[height=0.9em]{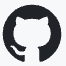}}~\textbf{Code:}~\url{https://github.com/RadiantCrystal/CRASE} \\[1pt]
        \end{center}
    \end{adjustwidth}%
}
\vspace{-0.7cm}

\title{Structurally-bounded Agentic Graph Exploration for \\Evidence-Grounded Scholarly DeepSearch}
\author{%
  Rima Hazra$^{\spadesuit\heartsuit}$\thanks{These authors contributed equally to this work.},
  Sayan Layek$^{\diamondsuit}$\footnotemark[1],
  Somnath Banerjee$^{\clubsuit}$, \\
  Soumen Chakrabarti$^{\dagger}$,
  \textbf{Animesh Mukherjee}$^{\diamondsuit}$\\
  }
  \affiliations{
  $^{\spadesuit}$National University of Singapore,
  $^{\heartsuit}$TCG CREST, \\
  $^{\diamondsuit}$Indian Institute of Technology Kharagpur,
  $^{\clubsuit}$Singapore Institute of Technology, \\
  $^{\dagger}$Indian Institute of Technology Bombay\\
  }
\usepackage{etoolbox}

\begin{document}

\maketitle

\begin{abstract}
We present \ours{}, a bounded and inspectable alternative to deep research agents for scholarly search. Instead of an open-ended search loop, \ours{} queries a search engine once for seed papers, expands them along their 1.5-hop citation neighborhood, prunes citation edges whose claims lack entailment support, and ranks the remaining papers with a recency-aware random walk. This makes the candidate set, the reason each paper is kept, and the stopping condition explicit and fixed before inference. On LitSearch and one further benchmarks over a 500K-paper \textbf{arXiv} corpus, \ours{} outperforms deep research agents built on proprietary models by up to 3$\times$ recall@50 at roughly a third of the cost.
\end{abstract}

\opensource

\section{Introduction}

Deep research agents turn a single-turn language model into an autonomous \textit{search-and-synthesis} process: given a complex information need, the agent
decomposes it into \textit{sub-questions}, \textit{calls external tools}, \textit{reads what they return},
\textit{judges whether evidences suffice}, and \textit{searches again before composing a cited
report} \cite{nakano2021webgpt,openai2025deepresearch,li2025searcho1}. Built on
the reasoning-and-acting paradigm for tool-augmented models
\cite{yao2023react,schick2023toolformer}, it is effective wherever evidence must be gathered from many sources, from literature review to multi-document comparison. This same autonomy, however, is hard to oversee. A user sees the final report but little of what produced it, why the agent kept searching or stopped, which sources shaped
its decisions, whether an early mistake quietly redirected everything after it.
Three properties of the loop compound the problem. Exploration is open-ended, so
a query halts on an external budget rather than any condition the user can
inspect. Errors persist, because a weak source pulled in early sits in the
context of every later query with no way to take it back. Also, evidence is hard to
audit, since the final citations seldom say why a paper was kept. There is a cost too -- often dozens of tool calls, the prompt growing at each step but the harder
problem is that open-ended exploration, opaque stopping, and compounding errors
leave these agents difficult to monitor and control. In scholarly discovery this
bites hardest: a report is only as reliable as the evidence behind it. We therefore ask:
\begin{researchquestion}
\emph{Can scholarly discovery be performed through a structurally bounded and
evidence-inspectable process without sacrificing retrieval effectiveness?}
\end{researchquestion}
Our starting point is simple: a generic agent spends its whole trajectory
rebuilding, one search at a time, a map of relevance the scientific literature
has already drawn, \textit{the citation network}
\cite{garfield1972citation,small1973cocitation}. So rather than rebuild that map,
we walk it. One seeded search replaces open-ended querying, the graph bounds
which papers are in play without asking the model to decide, and its edges leave
a visible trail of how they connect. This mirrors the AI control view of safe
autonomy like place a trusted, auditable component around an untrusted but capable
one, so guarantees come from structure rather than the model's own judgment of
when it has done enough. What enables the trade is a change of objective: we ask
the system not to synthesize an answer but to return a ranked set of relevant
papers, turning open-ended search into a process whose boundaries, steps, and
stopping point are open to inspection.

We instantiate this in \ours{} (\ourslong{}), which resolves a query in three
stages. It derives a research plan and issues one search call per sub-plan to
Semantic Scholar \cite{kinney2023semanticscholar}, keeping the top results as
seeds -- \textit{the only stage that touches an external service}. It expands the seeds to
their 1.5-hop citation neighborhood, \textit{the seeds}, \textit{their references} and \textit{citing
papers}, and every edge among them to form an \emph{\textbf{evidence graph}}. Raw citations are
often perfunctory, tangential, or purely methodological
\cite{moravcsik1975some,jurgens2018measuring}, and citation counts follow a
rich-get-richer dynamic \cite{barabasi1999emergence}: a heavily cited paper draws
edges through sheer popularity, so a naive walk pools relevance on such hubs and
drifts toward older work regardless of how well either grounds the query. \ours{}
counters both, scoring every edge by how strongly the cited paper's atomic claims
ground the citing paper's, so a popular but tangential hub is down-weighted or
pruned rather than rewarded for its degree, then ranking what remains with a
recency-aware Personalized PageRank \cite{page1999pagerank,haveliwala2002topic} or SALSA~\cite{Lempel2000TheSA} whose
weights balance this coverage affinity against an age decay. This claim-level reweighting is the central novelty, and it splits \ours{} into a
trusted part and a checkable one. The graph construction, pruning rule, and
ranking walk are fixed and deterministic, with no free choice left to the model;
the model supplies only local judgments, \textbf{the plan}, \textbf{the claims}, and \textbf{the per-edge
entailment scores}, each verifiable on its own. So the boundary and stopping
point are fixed rather than model-chosen, the edge weights record why each paper
was kept or cut, and the model drives external search only at seed construction.
\ours{} does not make its judgments correct, but it makes the search space,
evidence, and stopping condition auditable and while we study scholarly
discovery, the design is not specific to it.

\noindent We evaluate \ours{} on a controlled corpus of roughly 500K \textbf{arXiv} papers spanning
AI, ML, NLP, and computer vision (\textit{January} 2016 -- \textit{July} 2026), using LitSearch
\cite{ajith-etal-2024-litsearch} and one further test set restricted to this corpus.
\textbf{Our primary contributions are threefold}.
\begin{enumerate}[leftmargin=*, topsep=-2pt]\setlength{\itemsep}{1pt}\setlength{\parskip}{0pt}\setlength{\parsep}{0pt}

\item We identify critical challenges 
to scholarly deep research:
\textit{open-ended exploration}, \textit{model-controlled termination}, \textit{compounding errors}, and \textit{limited visibility into evidence selection}. In the spirit of AI control,
we \emph{recast} deep research as bounded discovery, where a trusted citation-graph substrate
tempers an untrusted model's judgments.

\item We introduce \emph{coverage affinity}, a claim-level entailment weighting of citation edges estimating how much a cited paper grounds the citing paper's claims, turning a bibliographic graph into an inspectable evidence graph that offsets both the popularity and recency biases of raw citation topology.

\item We rank papers by a recency-aware random walk over the pruned graph and show on LitSearch and one further benchmark that \ours{} matches or improves scholarly retrieval while substantially cutting the tool-call and token cost of generic deep research agents.
\end{enumerate}

\section{Related work}

\noindent \textbf{\textit{Agentic search and retrieval-augmented generation}}: Tool-augmented models interleave reasoning with search \cite{yao2023react,shinn2023reflexion}, learn tool use \cite{schick2023toolformer}, browse with human feedback \cite{nakano2021webgpt}, or combine retrieval with long-form reasoning and RL-trained search policies \cite{li2025searcho1,jin2025searchr1}; commercial deep research systems further extend this loop to report generation \cite{openai2025deepresearch} and dedicated benchmarks \cite{du2026deepresearch}. RAG methods similarly retrieve within generation \cite{lewis2020rag}, triggered by uncertainty \cite{jiang2023flare}, reasoning steps \cite{trivedi2023ircot}, or self-critique \cite{asai2023selfrag}. These approaches rely on repeated search, model-driven stopping, and largely irreversible retrieval trajectories; \ours{} instead performs a single seeded search and subsequently explores relations among retrieved papers.\\
\noindent \textbf{\textit{Scholarly retrieval and citation-graph ranking}}: Scientific search spans dedicated infrastructure \cite{kinney2023semanticscholar}, benchmarks \cite{ajith-etal-2024-litsearch}, citation recommenders \cite{farber2020citation}, and citation-aware embeddings \cite{cohan-etal-2020-specter,singh-etal-2023-scirepeval}; related systems interleave search with citation expansion \cite{he2025pasa} or combine retrieval with synthesis \cite{asai2024openscholar}. Citation-based relevance has long supported citation analysis \cite{garfield1972citation}, co-citation \cite{small1973cocitation}, bibliographic coupling \cite{kessler1963bibliographic}, and PageRank-based ranking \cite{page1999pagerank,haveliwala2002topic,jeh2003scaling}. However, raw citation graphs favor older papers through preferential attachment \cite{barabasi1999emergence} and conflate substantive with perfunctory citations \cite{moravcsik1975some}, motivating age-aware weighting and citation-intent modeling \cite{teufel2006automatic,jurgens2018measuring,cohan2019structural}. Instead of learning a step-wise expansion policy, \ours{} fixes a 1.5-hop neighborhood and weights edges by claim-level entailment \cite{bowman2015large,thorne2018fever,wadden2020scifact}, yielding an auditable evidence graph.\\
\noindent \textbf{\textit{Oversight and control of autonomous agents:}}
Scalable oversight makes model reasoning more supervisable by rewarding intermediate steps \cite{lightman2023verify} or testing whether weaker models can oversee stronger ones \cite{burns2024weaktostrong,bowman2022scalable}. AI control instead assumes the capable model may be unreliable and uses a trusted, auditable protocol to constrain it \cite{greenblatt2024aicontrol, rasheed2026fluentverifiableclaimlevelauditability}. A complementary line instead makes the model itself more reliable at inference time, through decoding-time or test-time alignment~\cite{banerjee2024safeinfercontextadaptivedecoding, banerjee2025prosocialalignpreferenceconditionedtest}, rather than constraining it externally. Deep research agents extend this challenge to multi-step evidence gathering, where these methods offer limited structure. \ours{} addresses this gap by separating the agent into a trusted substrate, fixed graph construction, coverage-affinity pruning, and deterministic ranking.

\section{Problem formulation}
A conventional deep research system follows a model-driven search loop. Starting from a query $q$, the model repeatedly invokes a search tool, inspects the retrieved evidence, updates its context, and decides whether another search is needed. Let $T$ denote the number of tool calls before termination. Since $T$ depends on the model's intermediate judgments, neither the extent of the explored literature nor the stopping point is known in advance. Moreover, the model itself determines which retrieved evidence influences later search decisions, making both evidence selection and termination difficult to audit. Consequently, it is difficult to understand why a particular paper was retrieved, why another was discarded, or why the search terminated at a particular point. We formulate the problem with the goal of making these decisions more explicit and structured. Rather than allowing the agent to perform an open-ended sequence of searches, we first define a bounded candidate space and then rank papers within this space. Given the query $q$, the objective is to return a ranked set $P_q$ containing at most $k$ relevant papers. This formulation makes the set of candidate papers, the evidence used for ranking, and the stopping condition directly observable. We represent a scientific corpus $\mathcal{C}$ as a directed citation graph. For each paper $p$, let $R_p$ denote its references and $C_p$ denote the papers that cite it. We define its citation neighborhood as $N(p)=R_p \cup C_p$.
Each paper is additionally represented by its textual content $t(p)$, consisting of its title, abstract, and introduction. Thus, a candidate paper can be evaluated using both its textual relevance to the query and its position within the citation structure.\\
\noindent Given $q$, we first retrieve a small set of seed papers $S \subseteq \mathcal{C}$ (see the section on seed construction). We then expand each seed through its citation neighborhood to obtain $V = S \cup \bigcup_{p \in S} N(p).$
Within this node set, we retain all citation links between the retrieved papers $E = {(u,v)\in E_{\mathcal{C}} : u,v\in V}$.
Although the nodes are obtained through a one-hop expansion, it also preserves citation relations among the retrieved papers themselves. We refer to this enriched structure as a \emph{1.5-hop neighborhood}. These additional relations provide evidence about how candidate papers are connected to the seed papers and to one another, which is subsequently used during ranking.
An important consequence of this construction is that the exploration space is fixed once $S$ is selected. Hence, the agent does not repeatedly decide whether to expand the search further. Similarly, evidence selection is performed over an explicit graph whose nodes, edges, and ranking signals can be inspected. The stopping condition is also fixed: the system ranks papers within this bounded graph and returns the top-$k$ candidates. The two controlling parameters are: $m$ (number of seed papers retained for each sub-query), and $k$ (number of papers returned after ranking).

\section{The \ours{} system}
\label{sec:overview}

\begin{figure*}[h]
\centering
\includegraphics[width=0.85\hsize]{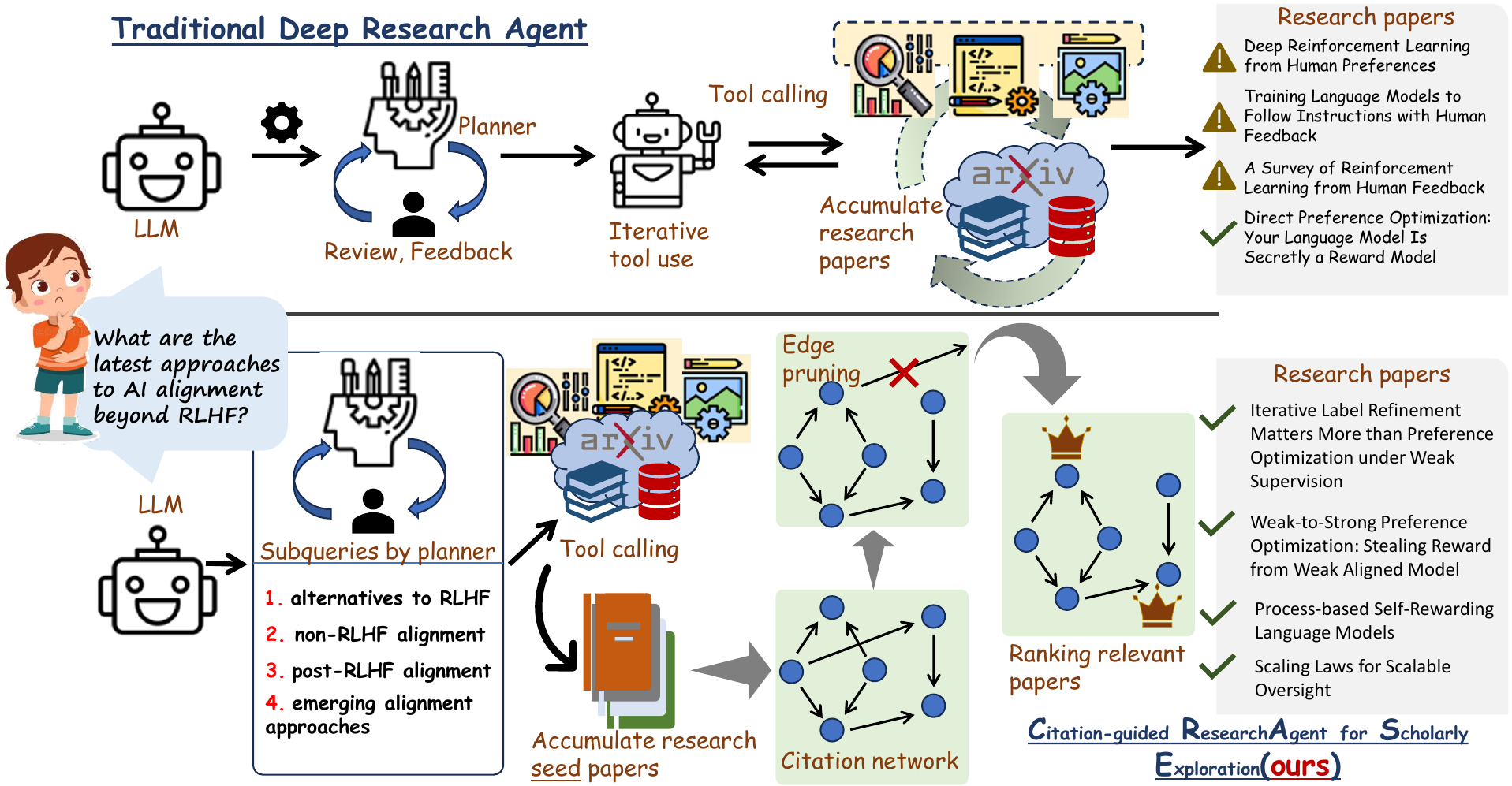}
\caption{\footnotesize Unlike deep research agents that repeatedly search until the model decides to stop, \ours{} performs a seeded search, constructs a 1.5-hop citation graph, prunes unsupported edges, and ranks papers with a recency-aware walk.}
\label{fig:overview}
\end{figure*}

\ours{} makes a single retrieval call to obtain a small seed set, then confines all further reasoning to the citation graph around those seeds, grounding discovery in the structure of the literature rather than in a sequence of model-issued searches. Three stages, shown in Figure~\ref{fig:overview}, carry a query from seeds to a ranked result.

\noindent \textbf{\underline{Stage 1 (\textit{Plan and seed construction}):}}
Given a query $q$, \ours{} decomposes it into a set of focused sub-queries $\mathscr{R}_q$. For each sub-query, it retrieves the top $m$ papers from an external index and forms their union as the seed set $\mathcal{S}$. This is the only stage that uses an external search service.

\noindent \textbf{\underline{Stage 2 (\textit{Guided graph expansion}):}}
\ours{} expands $\mathcal{S}$ to its $1.5$-hop citation neighborhood to construct an evidence graph $G=(V,E)$. Each citation edge is weighted by claim-level support between the citing and cited papers using a fine-tuned entailment model. Low-support edges are pruned, yielding a graph that reflects \emph{evidence-informed} relationships rather than citation structure alone.

\noindent \textbf{\underline{ Stage 3 (\textit{Ranking and output}):}}
Finally, \ours{} ranks the retained papers using a query-personalized, recency-aware random walk over $G$ and returns the top $k$ papers as $P_q$. The ranking module is general: we use standard personalized graph-ranking methods~\cite{page1999pagerank,Lempel2000TheSA}, but others can be plugged in.

\subsection{Plan and seed construction}
\label{subsec:seedcons}
The first stage of \ours{} converts a query into a small set of seed papers from which the evidence graph is expanded. It consists of two steps -- constructing a research plan and retrieving seeds for each plan element.

\noindent \textit{\underline{\smash{Research plan construction.}}}
Given a query $q$, we use an LLM to construct a research plan
$\mathscr{R}_q=\{\rho_1,\ldots,\rho_L\}$, where each $\rho_\ell$ is a concise keyword-style sub-query covering one facet of $q$. This decomposition provides multiple lexical entry points to the same information need, which is useful when relevant papers use different terminology. The prompt constrains each sub-query to preserve the technical requirements of $q$ without introducing new tasks, methods, datasets, or domains. We further require keyword-style queries to better match title- and abstract-based academic search. The full prompt is provided in supp. mat.

\noindent \textit{\underline{\smash{Seed construction.}}}
For each sub-query $\rho_\ell$, we issue a single Semantic Scholar search and obtain rankings by relevance and recency. We retain up to $m$ papers appearing among the top results of both rankings, favoring papers that are simultaneously relevant and recent. The final seed set is $\mathcal{S} = \bigcup_{\ell=1}^{L}\operatorname{Seeds}(\rho_\ell), |\operatorname{Seeds}(\rho_\ell)| \leq m$.
Since different sub-queries may retrieve the same papers, $|\mathcal{S}|$ can be smaller than $mL$. For each seed $p\in\mathcal{S}$, we then retrieve its references $R_p$ and citing papers $C_p$. These papers are not added directly to the seed set; instead, they define the local neighborhood used to construct the citation graph (worked example in supplementary material).

\subsection{Guided graph expansion}
\label{subsec:expansion}
The second stage expands the seed set into a weighted evidence graph whose edges reflect semantic support rather than mere presence of a citation. For each seed, we collect its references and citing papers (out- and in-links) and construct a directed graph $G=(V,E)$ over the resulting $1.5$-hop neighborhood, where $(u,v)\in E$ if $u$ cites $v$. Since a citation may be merely contextual or methodological, we assess whether the cited paper actually supports the main claims of the citing paper. To do so, we extract a small set of atomic claims from each paper, estimate claim-level support using a scientific entailment model, and aggregate these scores into a \emph{coverage affinity} $\Delta(u,v)$. We then use this affinity to prune unsupported edges and isolated nodes, and to weight the remaining graph for ranking.

\noindent \textbf{Claim extraction}:
We represent each paper $p\in V$ by a small set of atomic claims extracted from its title, abstract, and introduction using \emph{Qwen2.5-32B-Instruct}~\cite{qwen2025qwen25technicalreport} model:
\begin{equation}
  \mathcal{C}(p)=\{c^{p}_{1},\dots,c^{p}_{n_p}\},
  \qquad n_p=\lvert\mathcal{C}(p)\rvert\in\{4,5\},
\end{equation}
Each $c_i^p$ is a self-contained statement of a principal method, finding, or quantitative result. Restricting the representation to a few contribution-level claims filters background and related-work content that may introduce noise into subsequent grounding scores. The extraction prompt is provided in the supp mat. This step is one of the prime novelties introduced in \ours{}.

 
\noindent \textbf{Entailment model}:
The core operation in this stage is determining whether one scientific claim provides evidential grounding for another. Generic textual entailment models are trained largely on everyday natural language and may transfer poorly to dense, terminology-heavy scientific claims. We therefore fine-tune a dedicated entailment model for this setting. We use \textbf{MsciNLI}~\cite{sadat2024mscinlidiversebenchmarkscientific} and \textbf{SciNLI}~\cite{sadat-caragea-2022-scinli}, which provide premise--hypothesis pairs with entailment labels, and adapt them to our claim-grounding formulation. Specifically, we fine-tune \texttt{Qwen2.5-3B-Instruct} and evaluate it on the test sets of both datasets, where it achieves F1 scores of 0.8 and 0.83, respectively. We denote the resulting scoring function by $\operatorname{ent}(\cdot\mid\cdot)$ and use it below to score claim pairs along each citation edge. Full training and evaluation details are provided in the experiments section.


\noindent \textbf{Claim-level entailment:}
We apply the entailment model to each citation edge at the claim level. Consider a directed edge $(u,v)\in E$, where $u$ is the \emph{citing} paper and $v$ the \emph{cited} paper, with claim sets
$\mathcal{C}(u)=\{c^{u}_{1},\dots,c^{u}_{n_u}\}$ and
$\mathcal{C}(v)=\{c^{v}_{1},\dots,c^{v}_{n_v}\}$. We compute an entailment matrix $M^{(u,v)}\in[0,1]^{n_u\times n_v}$ whose entry
\begin{equation}
  M^{(u,v)}_{ij}=\operatorname{ent}\!\bigl(c^{u}_{i} \mid c^{v}_{j}\bigr)
\end{equation}
where $M^{(u,v)}_{ij}$ measures how strongly the reference claim $c^{v}{j}$ grounds the citing paper's claim $c^{u}{i}$. Since $n_u,n_v\leq 5$, the matrix contains at most $25$ entries. We obtain all entries in a single model call per edge, returned as JSON, rather than making $n_un_v$ separate calls, keeping the number of model calls linear in the number of edges.

\noindent \textbf{Claim support and coverage affinity:}
We next determine how strongly a reference $v$ grounds the claims of the citing paper $u$. Given a support threshold $\theta\in[0,1]$, a claim $c^{u}_{i}$ is considered supported by $v$ if at least one claim of $v$ entails it above this threshold. We define
\begin{equation}
  s_i(u,v)=\mathbf{1}\!\left[\,\max_{1\le j\le n_v} M^{(u,v)}_{ij}\ge\theta\,\right],
  \label{eq:support}
\end{equation}
where the row-wise maximum allows any single claim of $v$ to provide sufficient grounding for $c_i^u$. We then measure the \emph{coverage affinity} of the citation edge $(u,v)$ as the fraction of claims in $u$ supported by $v$:
\begin{equation}
  \Delta(u,v) =
  \frac{1}{n_u}\sum_{i=1}^{n_u} s_i(u,v)
  \in[0,1].
  \label{eq:delta}
\end{equation}
Thus, $\Delta(u,v)=1$ when $v$ grounds every claim of $u$ and $0$ when it grounds none. Since $\Delta(u,v)$ depends only on the entailment matrix $M^{(u,v)}$, each edge weight is computed independently of other references and is invariant to processing order. Multiple references may therefore receive credit for grounding the same claim.

\noindent \textbf{Edge and node pruning:}
We prune every edge with $\Delta(u,v)\le\tau$, using $\tau=0$ as a conservative default, so that
we discard an edge exactly when $v$ grounds no claim of $u$. We then remove any
node left without incident edges, since such a node can lie on no evidential path
and cannot contribute to ranking. The retained affinity $\Delta(u,v)$ is exactly
the value used in the ranking weight
\begin{equation}
  w(u\to v)=\Delta(u,v)\cdot\operatorname{age}(v)^{-\beta},
  \label{eq:weight}
\end{equation}
which couples semantic coverage with recency~\cite{Hu2021} and carries the pruned, reweighted
graph into the ranking stage.

\subsection{Source ranking on the evidence graph}
\label{sec:ranking}
The final stage of \ours{} produces the ranked result set $P_q$ from the
weighted evidence graph. Guided graph expansion 
yields a directed graph $G=(V,E)$ in which each edge $u\to v$ carries the weight
$w(u\to v)$ of Eq.~\ref{eq:weight}, coupling the coverage affinity $\Delta(u,v)$
with the recency term $\operatorname{age}(v)^{-\beta}$. The ranking stage is modular. \ours{} operates on the resulting weighted graph and can therefore be paired with different graph-based ranking algorithms. In our experiments, we consider
Personalized PageRank (PPR) and SALSA, and use PPR as the primary ranking mechanism. 
PPR propagates relevance from the seed papers through the weighted edges of $G$. Normalizing the outgoing edge weights gives the transition probability $P(u\to v)=\frac{w(u\to v)}{\sum_{v'} w(u\to v')}$ which forms the row-stochastic matrix $\mathbf{W}$. For any dangling node, we replace its row with the personalization distribution to keep $\mathbf{W}$ stochastic. We anchor the walk to the query through a \emph{personalization vector} $\mathbf{p}$, a distribution over $V$ concentrated on the seed set $\mathcal{S}$. Given $\mathbf{p}$ and a \emph{teleportation probability} $\alpha\in(0,1)$, the PPR score vector $\boldsymbol{\pi}$ satisfies the
fixed-point equation.
\begin{equation}
  \pi(v)=(1-\alpha)\sum_{u\to v}\pi(u)\,P(u\to v)+\alpha\,p(v),
  \label{eq:ppr-fixed-point}
\end{equation}
with closed-form solution
\begin{equation}
  \boldsymbol{\pi}=\alpha\bigl(\mathbf{I}-(1-\alpha)\mathbf{W}^{\!\top}\bigr)^{-1}\mathbf{p}.
  \label{eq:ppr-closed-form}
\end{equation}
At each step, the surfer follows an outgoing edge with probability $1-\alpha$ or
restarts from the seed distribution $\mathbf{p}$ with probability $\alpha$.
Because the transition probabilities derive from $w(u\to v)$, the walk favors
targets that are both well grounded in the seeds and recent. The stationary score
$\pi(v)$ thus measures the relevance of $v$ to the seeds, and hence to $q$. We
sort $V$ by $\boldsymbol{\pi}$ in descending order and return the top $k$ papers
as $P_q$.

\section{Datasets}
\noindent \textbf{Collected corpus: } Our corpus $\mathscr{C}$ comprises roughly $500\mathrm{K}$ \textbf{arXiv} papers submitted
between \textit{January} 2016 and \textit{July} 2026, spanning eight subject categories across
Artificial Intelligence, Machine Learning, NLP, and Computer Vision -- \cat{cs.AI},
\cat{cs.LG}, \cat{cs.CL}, \cat{cs.NE}, \cat{cs.CV}, \cat{cs.IR}, \cat{cs.MA}, and
\cat{stat.ML} -- where a paper is included if annotated with at least one; full
statistics are in the supplementary material. For each paper we collect its title,
publication year, abstract, and authors, and parse the introduction from the PDF
with GROBID~\cite{grobid}. The title, abstract, and introduction form the textual
representation $t(p)$ used throughout \ours{}, and the publication year supplies
the $\operatorname{age}$ in Eq.~\ref{eq:weight}.


\begin{table*}[h]
\centering
\setlength{\tabcolsep}{4.2pt}
\renewcommand{\arraystretch}{1.28}
\resizebox{0.80\textwidth}{!}{%
\begin{tabular}{ll cccccc | cccccc}
\toprule
\toprule
\multirow{2}{*}{\textbf{Dataset}} & \multirow{2}{*}{\textbf{Method}}
& \multicolumn{6}{c|}{\cellcolor{grouprbg}\textcolor{white}{\textbf{Recall@K}}}
& \multicolumn{6}{c}{\cellcolor{groupmbg}\textcolor{white}{\textbf{MAP@K}}} \\
\cmidrule(lr){3-8}\cmidrule(lr){9-14}
& & @5 & @10 & @20 & @30 & @40 & @50 & @5 & @10 & @20 & @30 & @40 & @50 \\
\midrule
\multirow{4}{*}{\textbf{\textbf{ACL}}}
 & \gpt{}           & \snd{0.0335} & 0.0696 & \snd{0.1289} & 0.1572 & 0.1881 & 0.2294 & 0.0107 & 0.0170 & 0.0234 & 0.0253 & 0.0267 & 0.0279 \\
 & \claude{}           & 0.0309 & 0.0696 & 0.1160 & 0.1366 & 0.1804 & 0.2294 & 0.0085 & 0.0150 & 0.0196 & 0.0203 & 0.0223 & 0.0242 \\
 & \textbf{\ours{}-Llama-PPR} & \best{0.0730} & \best{0.1236} & \best{0.1882} & \best{0.2584} & \best{0.3258} & \best{0.3848} & \best{0.0453} & \best{0.0530} & \best{0.0602} & \best{0.0656} & \best{0.0697} & \best{0.0730} \\
 & \spector{}  & 0.0284 & \snd{0.0747} & 0.1263 & \snd{0.1830} & \snd{0.2242} & \snd{0.2629} & \snd{0.0170} & \snd{0.0248} & \snd{0.0303} & \snd{0.0341} & \snd{0.0366} & \snd{0.0380} \\
\midrule
\multirow{4}{*}{\textbf{\textbf{ICLR}}}
 & \gpt{}            & 0.0183 & \snd{0.0427} & 0.0610 & 0.0732 & 0.1098 & 0.1220 & 0.0082 & 0.0106 & 0.0119 & 0.0123 & 0.0150 & 0.0155 \\
 & \claude{}             & 0.0183 & \snd{0.0427} & 0.0549 & 0.0671 & 0.1037 & 0.1220 & 0.0040 & 0.0074 & 0.0083 & 0.0094 & 0.0121 & 0.0126 \\
 & \textbf{\ours{}-Llama-PPR} & \best{0.1037} & \best{0.1402} & \best{0.2256} & \best{0.2866} & \best{0.3293} & \best{0.3720} & \best{0.0857} & \best{0.0903} & \best{0.0995} & \best{0.1044} & \best{0.1061} & \best{0.1076} \\
 & \spector{}  & \snd{0.0305} & \snd{0.0427} & \snd{0.0854} & \snd{0.1463} & \snd{0.1829} & \snd{0.2012} & \snd{0.0245} & \snd{0.0269} & \snd{0.0295} & \snd{0.0329} & \snd{0.0352} & \snd{0.0361} \\
\midrule
\multirow{4}{*}{\textbf{\textbf{arXiv}}}
 & \gpt{}           & 0.0491 & 0.0954 & 0.1811 & 0.2351 & 0.2793 & 0.3165 & 0.0270 & 0.0451 & 0.0738 & 0.0868 & 0.0935 & 0.0991 \\
 & \claude{}            & 0.0477 & 0.0940 & 0.1733 & 0.2246 & 0.2730 & 0.3165 & 0.0267 & 0.0454 & 0.0726 & 0.0845 & 0.0923 & 0.0987 \\
 & \textbf{\ours{}-Llama-PPR} & \snd{0.0533} & \best{0.1168} & \best{0.1988} & \best{0.2585} & \best{0.3102} & \best{0.3572} & \best{0.0541} & \snd{0.0635} & \snd{0.0899} & \best{0.1193} & \best{0.1303} & \snd{0.1168} \\
 & \spector{}  & \best{0.0674} & \snd{0.1151} & \snd{0.1916} & \snd{0.2533} & \snd{0.3018} & \snd{0.3446} & \snd{0.0494} & \best{0.0762} & \best{0.1040} & \snd{0.1183} & \snd{0.1269} & \best{0.1331} \\
\bottomrule
\bottomrule
\end{tabular}%
}
\caption{\footnotesize\textbf{Main results.} Recall@K and MAP@K across the three datasets and retrieval setups. \textbf{Best} results bold-colored; \underline{second-best} underlined. We observe that \ours{}-Llama-PPR consistently performs best on \textbf{ACL} and \textbf{ICLR} and remains competitive or best across most settings on \textbf{arXiv}.}
\label{tab:mainresults}
\vspace{-0.2cm}
\end{table*}

\noindent \textbf{Test datasets}: We evaluate \ours{} on three test sets that
differ mainly in how their queries and ground-truth papers are obtained. Each
query carries a set of relevant papers as ground-truth. We restrict both
queries and ground truth to $\mathscr{C}$ thus evaluating all methods against the
same candidate pool.


\noindent \textbf{LitSearch: } LitSearch~\cite{ajith-etal-2024-litsearch} pairs
literature-search queries with their relevant ground-truth papers; we use its
manual \textbf{\textbf{ACL}} and \textbf{\textbf{ICLR}} splits (155 and 91 queries). Restricting
each query's ground truth to $\mathscr{C}$ and dropping queries left with none
yields $114$ \textbf{\textbf{ACL}} and $60$ \textbf{\textbf{ICLR}} queries.

\noindent \textbf{\textbf{arXiv} set: } Our second set is built from $98$ papers submitted
to the eight categories of $\mathscr{C}$ between November 2025 and February 2026.
For each, we form a query from its title and abstract (prompt in the supplement)
and take its in-corpus reference list as ground-truth; we strip the related-work
section and references before generating the query, since these would reveal the
ground-truth. This set of papers sit at the end of the corpus span measuring retrieval on very recent work.

\section{Evaluation metrics}
\label{sec:metrics}
We measure retrieval quality with two standard metrics, Recall@$k$ and MAP@$k$,
aggregated over the queries $\mathcal{Q}$ of a dataset. For a query $q$, let
$\mathrm{Rel}(q)$ denote its set of ground-truth relevant papers and let
$P_q=(p_1,\dots,p_k)$ denote the top-$k$ papers returned by the method, ordered by decreasing score.

\noindent \textbf{Recall@$k$.}
Recall@$k$ is the proportion of relevant papers retrieved within the top $k$, pooled over all queries, $\mathrm{Recall}@k=\frac{\sum_{q\in\mathcal{Q}}\lvert P_q\cap\mathrm{Rel}(q)\rvert}{\sum_{q\in\mathcal{Q}}\lvert\mathrm{Rel}(q)\rvert}.$
Since scholarly search often extends beyond the top five results, we report multiple cutoffs: Recall@5 measures early-ranking quality, while Recall@20–Recall@50 capture broader evidence coverage.



\noindent \textbf{MAP@$k$.}
We also report mean average precision (MAP) at k, which rewards ranking relevant papers higher. Let $y_i=\mathbf{1}[\,p_i\in\mathrm{Rel}(q)\,]$
denote whether the paper at rank $i$ is relevant, and let $\mathrm{Prec}@i(q)$
denote the precision over the first $i$ papers. For a query $q$, the average
precision at $k$ is $\mathrm{AP}@k(q)=\frac{1}{\lvert\mathrm{Rel}(q)\rvert}
  \sum_{i=1}^{k} \mathrm{Prec}@i(q)\,y_i$ and MAP@$k$ is the mean over all queries, $\mathrm{MAP}@k=\frac{1}{\lvert\mathcal{Q}\rvert}\sum_{q\in\mathcal{Q}}
  \mathrm{AP}@k(q).$
\section{Experimental setup}
We configure \ours{} according to the three stages discussed in the earlier section. In Stage~1, \texttt{Qwen2.5-32B-Instruct} decomposes each query into focused sub-queries, and a single Semantic Scholar call per sub-query retrieves up to 10 seed papers. In Stage~2, claims are extracted using \texttt{Qwen2.5-32B-Instruct} and \texttt{Meta-Llama-3-70B-Instruct}. For claim entailment, we fine-tune \texttt{Qwen2.5-3B-Instruct} on MSciNLI and SciNLI to classify claim pairs into entailment, reasoning, contrasting, or neutral relations; we select the 3B model over its 1.5B counterpart based on its higher test accuracy. In Stage~3, the resulting weighted evidence graph is ranked using PPR, with teleportation probability $\alpha=0.15$, age-decay exponent $\beta=0.5$, and pruning threshold $\tau=0.75$. Full model, training, and ranking configurations are provided in the supplementary material.

\begin{table*}[h]
\centering
\setlength{\tabcolsep}{4.2pt}
\renewcommand{\arraystretch}{1.28}
\resizebox{0.80\textwidth}{!}{%
\begin{tabular}{ll cccccc | cccccc}
\toprule
\toprule
\multirow{2}{*}{\textbf{Dataset}} & \multirow{2}{*}{\textbf{Variant}}
& \multicolumn{6}{c|}{\cellcolor{grouprbg}\textcolor{white}{\textbf{Recall@K}}}
& \multicolumn{6}{c}{\cellcolor{groupmbg}\textcolor{white}{\textbf{MAP@K}}} \\
\cmidrule(lr){3-8}\cmidrule(lr){9-14}
& & @5 & @10 & @20 & @30 & @40 & @50 & @5 & @10 & @20 & @30 & @40 & @50 \\
\midrule
\multirow{4}{*}{\textbf{\textbf{ACL}}}
 & \ours{}-Qwen-SALSA   & \snd{0.0335} & \snd{0.0747} & 0.1263 & 0.1675 & 0.2113 & 0.2552 & \abest{0.0200} & \snd{0.0258} & \snd{0.0289} & 0.0317 & 0.0336 & 0.0359 \\
 & Recency            & \snd{0.0335} & 0.0619 & \snd{0.1521} & \snd{0.2165} & \snd{0.2655} & 0.3015 & 0.0160 & 0.0209 & 0.0281 & 0.0333 & 0.0360 & 0.0376 \\
 & \ours{}-Qwen-PPR     & 0.0309 & 0.0722 & \abest{0.1572} & \abest{0.2191} & \abest{0.2706} & \snd{0.3144} & 0.0148 & 0.0212 & 0.0284 & \snd{0.0336} & \snd{0.0362} & \snd{0.0382} \\
 & \ours{}-Llama-SALSA  & \abest{0.0393} & \abest{0.0787} & 0.1433 & 0.1938 & 0.2584 & \abest{0.3146} & \snd{0.0177} & \abest{0.0269} & \abest{0.0332} & \abest{0.0359} & \abest{0.0398} & \abest{0.0428} \\
\midrule
\multirow{4}{*}{\textbf{\textbf{ICLR}}}
 & \ours{}-Qwen-SALSA  & \snd{0.0549} & 0.0793 & 0.1280 & \snd{0.1768} & \snd{0.2073} & 0.2195 & 0.0197 & 0.0234 & 0.0280 & 0.0314 & 0.0325 & 0.0329 \\
 & Recency            & 0.0427 & \snd{0.1159} & \snd{0.2073} & \abest{0.2561} & \abest{0.3171} & \snd{0.3537} & 0.0235 & 0.0356 & 0.0469 & 0.0514 & 0.0545 & 0.0559 \\
 & \ours{}-Qwen-PPR     & 0.0488 & \abest{0.1220} & \abest{0.2195} & \abest{0.2561} & \abest{0.3171} & \abest{0.3659} & \snd{0.0245} & \snd{0.0372} & \snd{0.0497} & \snd{0.0531} & \snd{0.0561} & \snd{0.0580} \\
 & \ours{}-Llama-SALSA   & \abest{0.0671} & 0.0915 & 0.1341 & \snd{0.1768} & \snd{0.2073} & 0.2683 & \abest{0.0425} & \abest{0.0470} & \abest{0.0539} & \abest{0.0558} & \abest{0.0567} & \abest{0.0596} \\
\midrule
\multirow{4}{*}{\textbf{\textbf{arXiv}}}
 & \ours{}-Qwen-SALSA   & \abest{0.0716} & \abest{0.1312} & \snd{0.2035} & \snd{0.2625} & \snd{0.3081} & \snd{0.3523} & \snd{0.0479} & \snd{0.0723} & \snd{0.0961} & \snd{0.1081} & \snd{0.1156} & \snd{0.1210} \\
 & Recency            & 0.0519 & 0.1046 & 0.1944 & 0.2554 & 0.2968 & 0.3361 & 0.0380 & 0.0584 & 0.0832 & 0.0962 & 0.1032 & 0.1083 \\
 & \ours{}-Qwen-PPR     & 0.0519 & 0.1032 & 0.1902 & 0.2533 & 0.2947 & 0.3404 & 0.0384 & 0.0583 & 0.0821 & 0.0957 & 0.1028 & 0.1086 \\
 & \ours{}-Llama-SALSA  & \snd{0.0646} & \snd{0.1263} & \abest{0.2175} & \abest{0.2842} & \abest{0.3368} & \abest{0.3804} & \abest{0.0527} & \abest{0.0737} & \abest{0.1037} & \abest{0.1170} & \abest{0.1258} & \abest{0.1321} \\
\bottomrule
\bottomrule
\end{tabular}%
}
\caption{\footnotesize\textbf{Ablations.} Alternative claim extraction models (\textsc{Qwen}/\textsc{Llama}) and ranking strategies (\textsc{SALSA}/\textsc{PPR}) and the \textsc{Recency} baseline. \textbf{Best} per column is bold-colored; \underline{second-best} is underlined. We observe that \ours{}-Llama-SALSA gives the strongest overall performance, while \ours{}-Qwen-PPR is competitive in Recall@K.}
\label{tab:ablations}
\end{table*}
\subsection{Baselines}
\noindent \textbf{\gpt{}} - A GPT-based Deep Research baseline over the candidates
from guided graph expansion. Each paper node is represented by its title,
abstract, and introduction, embedded with \emph{BAAI/bge-base-en-v1.5}, and an
OpenAI model (\emph{o4-mini}) searches exclusively over these nodes through a
custom \emph{search graph papers} tool, ranking them by relevance to the query;
we keep the top-50.\\
\noindent \textbf{\claude{}} - The same setup, agentic framework, paper pool,
tools, and embedding model with the Anthropic model \emph{claude-sonnet-5} in
place of \emph{o4-mini}.\\
\noindent \textbf{\spector{}} - A scientific document embedder built on
SPECTER~\cite{cohan-etal-2020-specter} and
SciRepEval~\cite{singh-etal-2023-scirepeval}. We encode each candidate's title,
abstract, and introduction with its pretrained retrieval model, apply
DeepWalk~\cite{10.1145/2623330.2623732} to the expansion graph for structural
embeddings, and concatenate the two. Candidates are ranked by cosine similarity to seed papers.
\subsection{Other variants}
\noindent \textbf{\ours{} variants}: To test sensitivity to model and algorithmic
choices, we vary two components while holding all else like graph construction,
pruning, and the candidate pool fixed: the claim-extraction model
(\texttt{Qwen-32B} or \texttt{Llama-70B}) and the ranking algorithm over the
evidence graph (Personalized PageRank~\cite{page1999pagerank} or
SALSA~\cite{Lempel2000TheSA}). This yields four configurations, \ours{}-\{Qwen,
Llama\}-\{PPR, SALSA\}, each retaining the top-50 ranked papers.



\noindent \textbf{Recency}: This variant keeps \texttt{Qwen-32B} extraction and PPR ranking, changing only the age reference: paper age is measured relative to the query's publication year and not the current year.


\begin{table*}[t]
\centering
\setlength{\tabcolsep}{3pt}
\renewcommand{\arraystretch}{1.05}
\footnotesize
\resizebox{0.90\textwidth}{!}{%
\begin{tabular}{llcccccccccccc}
\toprule
\toprule
& & \multicolumn{6}{c}{\textbf{Recall@}$\bm{K}$} & \multicolumn{6}{c}{\textbf{MAP@}$\bm{K}$} \\
\cmidrule(lr){3-8} \cmidrule(lr){9-14}
\textbf{Dataset} & \textbf{Seeds} & \textbf{5} & \textbf{10} & \textbf{20} & \textbf{30} & \textbf{40} & \textbf{50} & \textbf{5} & \textbf{10} & \textbf{20} & \textbf{30} & \textbf{40} & \textbf{50} \\
\midrule
& Query yr.
& 0.0295 & 0.0516 & 0.1204 & 0.1622 & 0.2187 & 0.2654
& 0.0117 & 0.0158 & 0.0223 & 0.0253 & 0.0283 & 0.0311 \\
\rowcolor{oursrow}
\textbf{\textbf{ACL}} & Current yr.
& \textbf{0.0309} & \textbf{0.0722} & \textbf{0.1572} & \textbf{0.2191} & \textbf{0.2706} & \textbf{0.3144}
& \textbf{0.0148} & \textbf{0.0212} & \textbf{0.0284} & \textbf{0.0336} & \textbf{0.0362} & \textbf{0.0382} \\
& \scriptsize\color{gray}$\Delta$
& \scriptsize\color{up}$+.0014$ & \scriptsize\color{up}$+.0206$ & \scriptsize\color{up}$+.0368$ & \scriptsize\color{up}$+.0569$ & \scriptsize\color{up}$+.0519$ & \scriptsize\color{up}$+.0490$
& \scriptsize\color{up}$+.0031$ & \scriptsize\color{up}$+.0054$ & \scriptsize\color{up}$+.0061$ & \scriptsize\color{up}$+.0083$ & \scriptsize\color{up}$+.0079$ & \scriptsize\color{up}$+.0071$ \\
\midrule
& Query yr.
& \textbf{0.0606} & 0.1138 & 0.2036 & 0.2216 & 0.2515 & 0.2994
& \textbf{0.0278} & 0.0338 & 0.0441 & 0.0454 & 0.0463 & 0.0483 \\
\rowcolor{oursrow}
\textbf{\textbf{ICLR}} & Current yr.
& 0.0488 & \textbf{0.1220} & \textbf{0.2195} & \textbf{0.2561} & \textbf{0.3171} & \textbf{0.3659}
& 0.0245 & \textbf{0.0372} & \textbf{0.0497} & \textbf{0.0531} & \textbf{0.0561} & \textbf{0.0580} \\
& \scriptsize\color{gray}$\Delta$
& \scriptsize\color{down}$-.0118$ & \scriptsize\color{up}$+.0082$ & \scriptsize\color{up}$+.0159$ & \scriptsize\color{up}$+.0345$ & \scriptsize\color{up}$+.0656$ & \scriptsize\color{up}$+.0665$
& \scriptsize\color{down}$-.0033$ & \scriptsize\color{up}$+.0034$ & \scriptsize\color{up}$+.0056$ & \scriptsize\color{up}$+.0077$ & \scriptsize\color{up}$+.0098$ & \scriptsize\color{up}$+.0097$ \\
\bottomrule
\bottomrule
\end{tabular}
}
\caption{\footnotesize\textbf{Seed selection ablation.} Query-year vs.\ current-year seeds (shaded; used in main experiments); \textbf{bold} marks the best result and $\Delta$ the absolute change. Current-year seeds improve nearly all Recall@K and MAP@K scores, with the only drop at Recall@5 for ICLR.}
\label{tab:seedselectionrecallmap}
\end{table*}

\section{Results}
\label{sec:results}

Table~\ref{tab:mainresults} reports the retrieval performance on \textbf{ACL}, \textbf{ICLR}, and \textbf{arXiv}. We compare \ours{}-Llama-PPR with \gpt{}, \claude{}, and \spector{}. We report Recall$@$K and MAP$@$K for ($K\in\{5,10,20,30,40,50\}$). Overall, \ours{}-Llama-PPR gives the strongest results on \textbf{ACL} and \textbf{ICLR} datasets, while the results on \textbf{arXiv} dataset are more competitive.

\noindent \textbf{Recall$@$K}: On \textbf{ACL}, \ours{}-Llama-PPR achieves the best recall@K at all values of K. At recall$@$5, our method obtains 0.0730, compared to 0.0335 for the best baseline. The difference remains as K increases. At recall$@$20, recall$@$30, and recall$@$50, our method obtains 0.1882, 0.2584, and 0.3848, respectively, while the corresponding best baseline results are 0.1289, 0.1830, and 0.2629, respectively. Thus, the gain is not limited to any particular value of K. Similar to \textbf{ACL}, \ours{}-Llama-PPR achieves the best recall$@$K across all values of K on the \textbf{ICLR} dataset. In particular, recall$@$5 increases from 0.0305 for the best baseline to 0.1037 with our method. At recall$@$10 and recall$@$20, our method reaches 0.1402 and 0.2256, compared to 0.0427 and 0.0854 for the best baselines. The same trend continues at larger values of K, where \ours{}-Llama-PPR achieves a recall$@$50 of 0.3720, compared to 0.2012 for \spector{}. 
The \textbf{arXiv} dataset presents a different pattern. \spector{} performs best at recall$@$5 with 0.0674, while \ours{}-Llama-PPR obtains 0.0533. However, our method performs best for all remaining recall$@$K values. It improves recall$@$10 from 0.1151 to 0.1168 and recall@20 from 0.1916 to 0.1988. The difference becomes larger at higher (K): \ours{}-Llama-PPR obtains 0.3102 and 0.3572 at recall$@$40 and recall$@$50, compared to 0.3018 and 0.3446 for \spector{}. Hence, although \spector{} has an advantage for the recall$@$5 \textbf{arXiv} results, \ours{}-Llama-PPR recovers more relevant papers as the retrieval set grows.

\noindent \textbf{MAP$@$K}: The MAP results follow a similar trend on \textbf{ACL} and \textbf{ICLR} datasets. On \textbf{ACL} dataset, \ours{}-Llama-PPR achieves the best MAP$@$K, increasing MAP$@$5 from 0.0170 for the best baseline to 0.0453 and MAP$@$50 from 0.0380 to 0.0730. The difference is larger on \textbf{ICLR} dataset, where our method obtains MAP$@$5 of 0.0857 and MAP$@$50 of 0.1076, compared to 0.0245 and 0.0361 for \spector{}.
On \textbf{arXiv} dataset, we observe \ours{}-Llama-PPR performs best at MAP$@$5, MAP$@$30, and MAP$@$40, with scores of 0.0541, 0.1193, and 0.1303, respectively. \spector{} performs better at MAP$@$10, MAP$@$20, and MAP$@$50. This suggests that the two graph-based methods are closer on \textbf{arXiv} dataset, particularly in terms of the ranking of relevant papers.
\noindent Across the three datasets, \gpt{} and \claude{} generally remain behind the graph-based retrieval methods. The main difference is most clear on \textbf{ACL} and \textbf{ICLR} datasets, where \ours{}-Llama-PPR improves both recall and ranking quality over all baselines. On \textbf{arXiv}, the gap with \spector{} is smaller, but our method still obtains the best recall@K for five of the six retrieval depths. Overall, these results indicate that combining Llama-based representations with PPR provides stable retrieval performance across different scientific paper retrieval.

\noindent \textbf{Efficiency and interaction cost: } Table~\ref{tab:efficiency} compares retrieval utility with the interaction and computational cost of the agentic systems. The difference is particularly pronounced on the \textbf{ICLR} split. The ~\gpt{} and \claude{} agents require a median of 18 and 17 external tool calls per query, respectively, whereas \ours{} performs only five pre-specified external search calls during seed construction and performs no subsequent external search. \ours{} also substantially reduces total inference cost. It consumes approximately 235K tokens per query, compared to 620K for the \gpt{} and 560K for the \claude{} agent. Median end-to-end execution time decreases from 272 and 249 seconds, respectively, to 104 seconds with \ours{}. Under our cost accounting, the corresponding median per-query cost decreases from \$1.76 and \$2.06 to \$0.37. Importantly, these efficiency gains do not result from sacrificing retrieval quality. \ours{} achieves a recall@$50$ of 0.3659 on \textbf{ICLR} dataset, compared to 0.1220 for both proprietary-agent baselines. Thus, \ours{} achieves approximately $3\times$ higher recall@$50$ while operating with a substantially smaller and predetermined external interaction budget. This result demonstrates that structurally constraining agentic exploration can improve retrieval utility while simultaneously making resource consumption and external actions more predictable. 

\begin{table}[t] \centering \small \setlength{\tabcolsep}{1.6pt}  \resizebox{0.45\textwidth}{!}{\begin{tabular}{lrrrrr} \toprule \toprule \textbf{Method} & \textbf{Calls} $\downarrow$ & \textbf{Tokens} $\downarrow$ & \textbf{Time} $\downarrow$ & \textbf{Cost} $\downarrow$ & \textbf{R@50} $\uparrow$ \\ & & \textbf{(K)} & \textbf{(s)} & \textbf{(\$)} & \\ \midrule \gpt{} & 18 & 620 & 272 & 1.76 & 0.1220 \\ \claude{} & 17 & 560 & 249 & 2.06 & 0.1220 \\ ~\ours{}-Llama-PPR & \textbf{5} & \textbf{235} & \textbf{104} & \textbf{0.37} & \textbf{0.3659} \\ \bottomrule \bottomrule \end{tabular}}\caption{ \textbf{\footnotesize Retrieval utility and interaction cost on \textbf{ICLR}.} We report median per-query resource consumption together with recall@$50$. \ours{} achieves substantially higher retrieval recall while requiring fewer external interactions and lower inference cost. $\downarrow$ indicates lower is better and $\uparrow$ indicates higher is better.} \label{tab:efficiency}
\vspace{-0.3cm}
\end{table}

\begin{tcolorbox}[colback=blue!5, colframe=black!60]
\footnotesize 
\noindent \textbf{\underline{Are \ours{}'s evidence decisions human-auditable?}}
We evaluate whether the retain/prune decisions exposed by the evidence graph align with independent human judgments. On 50 citation edges (Table~\ref{tab:edge-audit}), \ours{} agrees with the majority expert judgment in \textbf{84.0\%} of cases. Experts validate \textbf{88.0\%} of retained edges as meaningful evidence or topical grounding and \textbf{80.0\%} of pruned edges as appropriate to discard. Human judgments are also consistent, with \textbf{68.0\%} unanimous agreement and Fleiss' $\kappa=\textbf{0.71}$. Thus, \ours{} exposes intermediate evidence decisions that largely align with human judgment and can be independently inspected before the final ranking.
\end{tcolorbox}
\begin{table}[t] \centering \small \setlength{\tabcolsep}{2pt}  \resizebox{0.29\textwidth}{!}{\begin{tabular}{lc} \toprule \toprule \textbf{Measure} & \textbf{Score} \\ \midrule \ours{}--human agreement & \textbf{84.0\%} \\ Retained-edge precision & \textbf{88.0\%} \\ Pruned-edge correctness & \textbf{80.0\%} \\ Unanimous human agreement & 68.0\% \\ Fleiss' $\kappa$ & \textbf{0.71} \\ \bottomrule \bottomrule \end{tabular}}\caption{ \textbf{\footnotesize Human validation of evidence-graph decisions.} Three domain experts independently assess 50 citation edges while being blinded to \ours{}'s decision. \ours{}'s retain/prune decisions are compared against the majority human judgment. } \label{tab:edge-audit}\end{table}

\noindent \textbf{Effect of design choices}: We further study the effect of the claim extraction models and ranking strategy in Table~\ref{tab:ablations}. Overall, \ours{}-Llama-SALSA provides the most consistent performance across datasets, particularly for MAP@K. On \textbf{ACL} dataset, it achieves the best MAP for K=10 through 50, while \ours{}-Qwen-PPR gives the best recall@K for K=20, 30, and 40. A similar pattern is observed on \textbf{ICLR}. \ours{}-Llama-SALSA achieves the best MAP at every K, whereas \ours{}-Qwen-PPR performs better for Recall at larger values of K. On \textbf{arXiv} dataset, \ours{}-Llama-SALSA achieves the best Recall for K=20 through 50 and the best MAP at every K. In contrast, \ours{}-Qwen-SALSA performs better at smaller recall cutoffs, obtaining the best recall$@$5 and recall$@$10. The Recency variant remains competitive on recall metric for \textbf{ICLR} dataset, but is generally weaker in MAP and on the other datasets. These results suggest that both the retrieval backbone and the ranking strategy affect performance, with the Llama-SALSA combination providing the most stable results across datasets and evaluation metrics.
\\
\noindent \textbf{Effect of seed selection strategy}: Table~\ref{tab:seedselectionrecallmap} compares two ways of picking seeds: using the year of the query, and using the current year. On \textbf{ACL}, seeding by the current year is better on every column, all six recall cutoffs and all six MAP cutoffs. For example, recall@50 rises from 0.2654 to 0.3144 and MAP@50 from 0.0311 to 0.0382. On \textbf{ICLR}, seeding by the current year is better at all but the two smallest cutoffs. Seeding by the year of the query is slightly ahead at recall@5 (0.0606 against 0.0488) and MAP@5 (0.0278 against 0.0245), but from recall@10 and MAP@10 onward the current-year strategy leads. Recall@50 goes from 0.2994 to 0.3659 and MAP@50 from 0.0483 to 0.0580.
On both datasets, the gap grows as $K$ increases, so the benefit of current-year seeding is largest at the deeper cutoffs. We therefore use current-year seed selection in \ours{}.
\begin{table}[t]
\tiny
\centering
\setlength{\tabcolsep}{8pt}
\renewcommand{\arraystretch}{1.05}
\footnotesize
\resizebox{0.35\textwidth}{!}{
\begin{tabular}{lcc}
\toprule
\toprule
\textbf{Model} & \textbf{MSciNLI} & \textbf{SciNLI} \\
\midrule
Qwen2.5-1.5B-Instruct & 0.75 & 0.77 \\
Qwen2.5-3B-Instruct   & \textbf{0.80} & \textbf{0.83} \\
\bottomrule
\bottomrule
\end{tabular}
}
\caption{\footnotesize F1 score on claim entailment. Best per dataset in \textbf{bold}. \emph{Qwen2.5-3B} consistently outperforms \emph{Qwen2.5-1.5B} on both datasets.}
\label{tab:model_accuracy}
\end{table}

\noindent \textbf{Claim entailment model result}:
Table~\ref{tab:model_accuracy} compares two Qwen2.5 models on MSciNLI and SciNLI under a two-class setting. Increasing the model size from ~\emph{1.5B} to~\emph{3B} improves F1 score by $\sim 5$ points on both MSciNLI ($75\%\rightarrow80\%$) and SciNLI ($77\%\rightarrow83\%$). Both models achieve higher accuracy on SciNLI than on MSciNLI. Based on these results, we use \texttt{Qwen2.5-3B-Instruct} for claim entailment in \ours{}.\\
\textbf{Complexity analysis}: The cost of \ours{} is fixed by the size of the evidence graph, not by a
model-chosen trajectory. With $|S|\le mL$ seeds and maximum seed degree
$d_{\max}$, the graph has $|V| = O(mL\,d_{\max})$ nodes and $|E| = O(|V|\,\bar d)$
edges, both determined by $S$ \emph{before} any inference. 
\if{0}The end-to-end cost is
\begin{equation}
O\big(
\underbrace{|V|\,c_{\mathrm{claim}} + |E|\,c_{\mathrm{ent}}}_{\text{graph weighting}}
+ \underbrace{\tfrac{1}{\alpha}\log\tfrac{1}{\epsilon}\,(|V|+|E|)}_{\text{ranking}}
\big),
\label{eq:complexity}
\end{equation}\fi
The complexity is dominated by the $|E|$ entailment calls, and the total number of model
invocations is exactly $1 + |V| + |E|$, \emph{known before inference and
independent of intermediate outputs} (see supplementary material for full derivations). A generic search-and-synthesis agent, by
contrast, appends each $O(\ell)$ observation to its context over
$T_{\mathrm{tool}}$ model-terminated tool calls, giving cumulative cost
$O(T_{\mathrm{tool}}^2\,\ell)$: quadratic in a trajectory length that is
unbounded a priori. \ours{} is instead \emph{linear} in a structurally fixed
graph, with constant-size prompts and a stopping condition set before inference.

\section{Conclusion}

We introduced \ours{}, a structurally bounded approach to scholarly deep research that replaces open-ended search with citation-guided exploration and claim-level evidence filtering. \ours{} improves retrieval over agentic baselines while using fewer external calls, tokens, time, and cost, and its evidence decisions largely align with human judgments. These results suggest that constraining agentic search around explicit, inspectable evidence can improve both retrieval effectiveness and auditability.


\bibliography{aaai2027}

@inproceedings{10.1145/2623330.2623732,
author = {Perozzi, Bryan and Al-Rfou, Rami and Skiena, Steven},
title = {DeepWalk: online learning of social representations},
year = {2014},
isbn = {9781450329569},
publisher = {Association for Computing Machinery},
address = {New York, NY, USA},
url = {https://doi.org/10.1145/2623330.2623732},
doi = {10.1145/2623330.2623732},
booktitle = {Proceedings of the 20th ACM SIGKDD International Conference on Knowledge Discovery and Data Mining},
pages = {701–710},
numpages = {10},
location = {New York, New York, USA},
series = {KDD '14}
}

@misc{rasheed2026fluentverifiableclaimlevelauditability,
      title={From Fluent to Verifiable: Claim-Level Auditability for Deep Research Agents}, 
      author={Razeen A Rasheed and Somnath Banerjee and Animesh Mukherjee and Rima Hazra},
      year={2026},
      eprint={2602.13855},
      archivePrefix={arXiv},
      primaryClass={cs.AI},
      url={https://arxiv.org/abs/2602.13855}, 
}

@misc{banerjee2024safeinfercontextadaptivedecoding,
      title={SafeInfer: Context Adaptive Decoding Time Safety Alignment for Large Language Models}, 
      author={Somnath Banerjee and Sayan Layek and Soham Tripathy and Shanu Kumar and Animesh Mukherjee and Rima Hazra},
      year={2024},
      eprint={2406.12274},
      archivePrefix={arXiv},
      primaryClass={cs.CL},
      url={https://arxiv.org/abs/2406.12274}, 
}

@misc{banerjee2025prosocialalignpreferenceconditionedtest,
      title={ProSocialAlign: Preference Conditioned Test Time Alignment in Language Models}, 
      author={Somnath Banerjee and Sayan Layek and Sayantan Adak and Mykola Pechenizkiy and Animesh Mukherjee and Rima Hazra},
      year={2025},
      eprint={2512.06515},
      archivePrefix={arXiv},
      primaryClass={cs.CL},
      url={https://arxiv.org/abs/2512.06515}, 
}

@misc{qwen2025qwen25technicalreport,
      title={Qwen2.5 Technical Report}, 
      author={Qwen},
      year={2025},
      eprint={2412.15115},
      archivePrefix={arXiv},
      primaryClass={cs.CL},
      url={https://arxiv.org/abs/2412.15115}, 
}

@misc{sadat2024mscinlidiversebenchmarkscientific,
      title={MSciNLI: A Diverse Benchmark for Scientific Natural Language Inference}, 
      author={Mobashir Sadat and Cornelia Caragea},
      year={2024},
      eprint={2404.08066},
      archivePrefix={arXiv},
      primaryClass={cs.CL},
      url={https://arxiv.org/abs/2404.08066}, 
}

@inproceedings{sadat-caragea-2022-scinli,
    title = "{S}ci{NLI}: A Corpus for Natural Language Inference on Scientific Text",
    author = "Sadat, Mobashir  and
      Caragea, Cornelia",
    editor = "Muresan, Smaranda  and
      Nakov, Preslav  and
      Villavicencio, Aline",
    booktitle = "Proceedings of the 60th Annual Meeting of the Association for Computational Linguistics (Volume 1: Long Papers)",
    month = may,
    year = "2022",
    address = "Dublin, Ireland",
    publisher = "Association for Computational Linguistics",
    url = "https://aclanthology.org/2022.acl-long.511/",
    doi = "10.18653/v1/2022.acl-long.511",
    pages = "7399--7409"
}

@inproceedings{singh-etal-2023-scirepeval,
    title = "{S}ci{R}ep{E}val: A Multi-Format Benchmark for Scientific Document Representations",
    author = "Singh, Amanpreet  and
      D{'}Arcy, Mike  and
      Cohan, Arman  and
      Downey, Doug  and
      Feldman, Sergey",
    editor = "Bouamor, Houda  and
      Pino, Juan  and
      Bali, Kalika",
    booktitle = "Proceedings of the 2023 Conference on Empirical Methods in Natural Language Processing",
    month = dec,
    year = "2023",
    address = "Singapore",
    publisher = "Association for Computational Linguistics",
    url = "https://aclanthology.org/2023.emnlp-main.338/",
    doi = "10.18653/v1/2023.emnlp-main.338",
    pages = "5548--5566"
}

@inproceedings{ajith-etal-2024-litsearch,
    title = "{L}it{S}earch: A Retrieval Benchmark for Scientific Literature Search",
    author = "Ajith, Anirudh  and
      Xia, Mengzhou  and
      Chevalier, Alexis  and
      Goyal, Tanya  and
      Chen, Danqi  and
      Gao, Tianyu",
    editor = "Al-Onaizan, Yaser  and
      Bansal, Mohit  and
      Chen, Yun-Nung",
    booktitle = "Proceedings of the 2024 Conference on Empirical Methods in Natural Language Processing",
    month = nov,
    year = "2024",
    address = "Miami, Florida, USA",
    publisher = "Association for Computational Linguistics",
    url = "https://aclanthology.org/2024.emnlp-main.840/",
    doi = "10.18653/v1/2024.emnlp-main.840",
    pages = "15068--15083"
}

@inproceedings{cohan-etal-2020-specter,
    title = "{SPECTER}: Document-level Representation Learning using Citation-informed Transformers",
    author = "Cohan, Arman  and
      Feldman, Sergey  and
      Beltagy, Iz  and
      Downey, Doug  and
      Weld, Daniel",
    editor = "Jurafsky, Dan  and
      Chai, Joyce  and
      Schluter, Natalie  and
      Tetreault, Joel",
    booktitle = "Proceedings of the 58th Annual Meeting of the Association for Computational Linguistics",
    month = jul,
    year = "2020",
    address = "Online",
    publisher = "Association for Computational Linguistics",
    url = "https://aclanthology.org/2020.acl-main.207/",
    doi = "10.18653/v1/2020.acl-main.207",
    pages = "2270--2282"
}

@misc{grobid,
    title = {GROBID},
    howpublished = {\url{https://github.com/grobidOrg/grobid}},
    publisher = {GitHub},
    year = {2008--2026}
}

@article{nakano2021webgpt,
  title   = {WebGPT: Browser-Assisted Question-Answering with Human Feedback},
  author  = {Nakano, Reiichiro and Hilton, Jacob and Balaji, Suchir and Wu, Jeff and Ouyang, Long and Kim, Christina and Hesse, Christopher and Jain, Shantanu and Kosaraju, Vineet and Saunders, William and others},
  journal = {arXiv preprint arXiv:2112.09332},
  year    = {2021}
}

@inproceedings{yao2023react,
  title     = {ReAct: Synergizing Reasoning and Acting in Language Models},
  author    = {Yao, Shunyu and Zhao, Jeffrey and Yu, Dian and Du, Nan and Shafran, Izhak and Narasimhan, Karthik and Cao, Yuan},
  booktitle = {International Conference on Learning Representations (ICLR)},
  year      = {2023}
}

@inproceedings{schick2023toolformer,
  title     = {Toolformer: Language Models Can Teach Themselves to Use Tools},
  author    = {Schick, Timo and Dwivedi-Yu, Jane and Dess{\`\i}, Roberto and Raileanu, Roberta and Lomeli, Maria and Hambro, Eric and Zettlemoyer, Luke and Cancedda, Nicola and Scialom, Thomas},
  booktitle = {Advances in Neural Information Processing Systems (NeurIPS)},
  year      = {2023}
}

@inproceedings{shinn2023reflexion,
  title     = {Reflexion: Language Agents with Verbal Reinforcement Learning},
  author    = {Shinn, Noah and Cassano, Federico and Gopinath, Ashwin and Narasimhan, Karthik and Yao, Shunyu},
  booktitle = {Advances in Neural Information Processing Systems (NeurIPS)},
  year      = {2023}
}

@article{li2025searcho1,
  title   = {Search-o1: Agentic Search-Enhanced Large Reasoning Models},
  author  = {Li, Xiaoxi and Dong, Guanting and Jin, Jiajie and Zhang, Yuyao and Zhou, Yujia and Zhu, Yutao and Zhang, Peitian and Dou, Zhicheng},
  journal = {arXiv preprint arXiv:2501.05366},
  year    = {2025}
}

@article{jin2025searchr1,
  title   = {Search-R1: Training LLMs to Reason and Leverage Search Engines with Reinforcement Learning},
  author  = {Jin, Bowen and Zeng, Hansi and Yue, Zhenrui and Yoon, Jinsung and Arik, Sercan and Wang, Dong and Zamani, Hamed and Han, Jiawei},
  journal = {arXiv preprint arXiv:2503.09516},
  year    = {2025}
}

@misc{openai2025deepresearch,
  title        = {Introducing Deep Research},
  author       = {{OpenAI}},
  year         = {2025},
  howpublished = {\url{https://openai.com/index/introducing-deep-research/}}
}

@inproceedings{du2026deepresearch,
title={DeepResearch Bench: A Comprehensive Benchmark for Deep Research Agents},
author={Mingxuan Du and Benfeng Xu and Chiwei Zhu and Licheng Zhang and Xiaorui Wang and Zhendong Mao},
booktitle={The Fourteenth International Conference on Learning Representations},
year={2026},
url={https://openreview.net/forum?id=hQ0K2Hhq7H}
}

@inproceedings{lewis2020rag,
  title     = {Retrieval-Augmented Generation for Knowledge-Intensive NLP Tasks},
  author    = {Lewis, Patrick and Perez, Ethan and Piktus, Aleksandra and Petroni, Fabio and Karpukhin, Vladimir and Goyal, Naman and K{\"u}ttler, Heinrich and Lewis, Mike and Yih, Wen-tau and Rockt{\"a}schel, Tim and Riedel, Sebastian and Kiela, Douwe},
  booktitle = {Advances in Neural Information Processing Systems (NeurIPS)},
  year      = {2020}
}

@inproceedings{jiang2023flare,
  title     = {Active Retrieval Augmented Generation},
  author    = {Jiang, Zhengbao and Xu, Frank F. and Gao, Luyu and Sun, Zhiqing and Liu, Qian and Dwivedi-Yu, Jane and Yang, Yiming and Callan, Jamie and Neubig, Graham},
  booktitle = {Proceedings of the 2023 Conference on Empirical Methods in Natural Language Processing (EMNLP)},
  year      = {2023}
}

@inproceedings{trivedi2023ircot,
  title     = {Interleaving Retrieval with Chain-of-Thought Reasoning for Knowledge-Intensive Multi-Step Questions},
  author    = {Trivedi, Harsh and Balasubramanian, Niranjan and Khot, Tushar and Sabharwal, Ashish},
  booktitle = {Proceedings of the 61st Annual Meeting of the Association for Computational Linguistics (ACL)},
  year      = {2023}
}

@article{asai2023selfrag,
  title   = {Self-RAG: Learning to Retrieve, Generate, and Critique through Self-Reflection},
  author  = {Asai, Akari and Wu, Zeqiu and Wang, Yizhong and Sil, Avirup and Hajishirzi, Hannaneh},
  journal = {arXiv preprint arXiv:2310.11511},
  year    = {2023}
}

@inproceedings{he2025pasa,
  title     = {PaSa: An LLM Agent for Comprehensive Academic Paper Search},
  author    = {He, Yichen and Huang, Guanhua and Feng, Peiyuan and Lin, Yuan and Zhang, Yuchen and Li, Hang and E, Weinan},
  booktitle = {Proceedings of the 63rd Annual Meeting of the Association for Computational Linguistics (ACL)},
  pages     = {11663--11679},
  year      = {2025},
  address   = {Vienna, Austria},
  publisher = {Association for Computational Linguistics}
}

@article{asai2024openscholar,
  title   = {OpenScholar: Synthesizing Scientific Literature with Retrieval-Augmented LMs},
  author  = {Asai, Akari and He, Jacqueline and Shao, Rulin and Shi, Weijia and Singh, Amanpreet and Chang, Joseph Chee and Lo, Kyle and Soldaini, Luca and Feldman, Sergey and D'Arcy, Mike and others},
  journal = {arXiv preprint arXiv:2411.14199},
  year    = {2024}
}

@article{Hu2021,
   title={The aging effect in evolving scientific citation networks},
   volume={126},
   ISSN={1588-2861},
   url={http://dx.doi.org/10.1007/s11192-021-03929-8},
   DOI={10.1007/s11192-021-03929-8},
   number={5},
   journal={Scientometrics},
   publisher={Springer Science and Business Media LLC},
   author={Hu, Feng and Ma, Lin and Zhan, Xiu-Xiu and Zhou, Yinzuo and Liu, Chuang and Zhao, Haixing and Zhang, Zi-Ke},
   year={2021},
   month=Mar, pages={4297–4309} }

@article{kinney2023semanticscholar,
  title   = {The Semantic Scholar Open Data Platform},
  author  = {Kinney, Rodney and Anastasiades, Chloe and Authur, Russell and Beltagy, Iz and Bragg, Jonathan and Buraczynski, Alexandra and Cachola, Isabel and Candra, Stefan and Chandrasekhar, Yoganand and Cohan, Arman and others},
  journal = {arXiv preprint arXiv:2301.10140},
  year    = {2023}
}

@article{farber2020citation,
  title   = {Citation Recommendation: Approaches and Datasets},
  author  = {F{\"a}rber, Michael and Jatowt, Adam},
  journal = {International Journal on Digital Libraries},
  volume  = {21},
  number  = {4},
  pages   = {375--405},
  year    = {2020}
}

@article{garfield1972citation,
  title   = {Citation Analysis as a Tool in Journal Evaluation},
  author  = {Garfield, Eugene},
  journal = {Science},
  volume  = {178},
  number  = {4060},
  pages   = {471--479},
  year    = {1972}
}

@article{small1973cocitation,
  title   = {Co-citation in the Scientific Literature: A New Measure of the Relationship Between Two Documents},
  author  = {Small, Henry},
  journal = {Journal of the American Society for Information Science},
  volume  = {24},
  number  = {4},
  pages   = {265--269},
  year    = {1973}
}

@article{kessler1963bibliographic,
  title   = {Bibliographic Coupling Between Scientific Papers},
  author  = {Kessler, Maxwell Mirton},
  journal = {American Documentation},
  volume  = {14},
  number  = {1},
  pages   = {10--25},
  year    = {1963}
}

@techreport{page1999pagerank,
  title       = {The PageRank Citation Ranking: Bringing Order to the Web},
  author      = {Page, Lawrence and Brin, Sergey and Motwani, Rajeev and Winograd, Terry},
  institution = {Stanford InfoLab},
  year        = {1999}
}

@inproceedings{haveliwala2002topic,
  title     = {Topic-Sensitive PageRank},
  author    = {Haveliwala, Taher H.},
  booktitle = {Proceedings of the 11th International Conference on World Wide Web (WWW)},
  pages     = {517--526},
  year      = {2002}
}

@inproceedings{jeh2003scaling,
  title     = {Scaling Personalized Web Search},
  author    = {Jeh, Glen and Widom, Jennifer},
  booktitle = {Proceedings of the 12th International Conference on World Wide Web (WWW)},
  pages     = {271--279},
  year      = {2003}
}

@article{barabasi1999emergence,
  title   = {Emergence of Scaling in Random Networks},
  author  = {Barab{\'a}si, Albert-L{\'a}szl{\'o} and Albert, R{\'e}ka},
  journal = {Science},
  volume  = {286},
  number  = {5439},
  pages   = {509--512},
  year    = {1999}
}

@article{moravcsik1975some,
  title   = {Some Results on the Function and Quality of Citations},
  author  = {Moravcsik, Michael J. and Murugesan, Poovanalingam},
  journal = {Social Studies of Science},
  volume  = {5},
  number  = {1},
  pages   = {86--92},
  year    = {1975}
}

@inproceedings{teufel2006automatic,
  title     = {Automatic Classification of Citation Function},
  author    = {Teufel, Simone and Siddharthan, Advaith and Tidhar, Dan},
  booktitle = {Proceedings of the 2006 Conference on Empirical Methods in Natural Language Processing (EMNLP)},
  pages     = {103--110},
  year      = {2006}
}

@article{jurgens2018measuring,
  title   = {Measuring the Evolution of a Scientific Field through Citation Frames},
  author  = {Jurgens, David and Kumar, Srijan and Hoover, Raine and McFarland, Dan and Jurafsky, Dan},
  journal = {Transactions of the Association for Computational Linguistics},
  volume  = {6},
  pages   = {391--406},
  year    = {2018}
}

@inproceedings{cohan2019structural,
  title     = {Structural Scaffolds for Citation Intent Classification in Scientific Publications},
  author    = {Cohan, Arman and Ammar, Waleed and van Zuylen, Madeleine and Cady, Field},
  booktitle = {Proceedings of the 2019 Conference of the North American Chapter of the Association for Computational Linguistics (NAACL)},
  year      = {2019}
}

@inproceedings{bowman2015large,
  title     = {A Large Annotated Corpus for Learning Natural Language Inference},
  author    = {Bowman, Samuel R. and Angeli, Gabor and Potts, Christopher and Manning, Christopher D.},
  booktitle = {Proceedings of the 2015 Conference on Empirical Methods in Natural Language Processing (EMNLP)},
  year      = {2015}
}

@inproceedings{thorne2018fever,
  title     = {FEVER: A Large-scale Dataset for Fact Extraction and VERification},
  author    = {Thorne, James and Vlachos, Andreas and Christodoulopoulos, Christos and Mittal, Arpit},
  booktitle = {Proceedings of the 2018 Conference of the North American Chapter of the Association for Computational Linguistics (NAACL)},
  year      = {2018}
}

@inproceedings{wadden2020scifact,
  title     = {Fact or Fiction: Verifying Scientific Claims},
  author    = {Wadden, David and Lin, Shanchuan and Lo, Kyle and Wang, Lucy Lu and van Zuylen, Madeleine and Cohan, Arman and Hajishirzi, Hannaneh},
  booktitle = {Proceedings of the 2020 Conference on Empirical Methods in Natural Language Processing (EMNLP)},
  year      = {2020}
}

@article{Lempel2000TheSA,
author = {Lempel, R. and Moran, S.},
title = {SALSA: the stochastic approach for link-structure analysis},
year = {2001},
issue_date = {April 2001},
publisher = {Association for Computing Machinery},
address = {New York, NY, USA},
volume = {19},
number = {2},
issn = {1046-8188},
url = {https://doi.org/10.1145/382979.383041},
doi = {10.1145/382979.383041},
journal = {ACM Trans. Inf. Syst.},
month = apr,
pages = {131–160},
numpages = {30}
}

@inproceedings{lightman2023verify,
  title     = {Let's Verify Step by Step},
  author    = {Lightman, Hunter and Kosaraju, Vineet and Burda, Yura and
               Edwards, Harri and Baker, Bowen and Lee, Teddy and Leike, Jan and
               Schulman, John and Sutskever, Ilya and Cobbe, Karl},
  booktitle = {International Conference on Learning Representations (ICLR)},
  year      = {2024},
  note      = {arXiv:2305.20050}
}

@inproceedings{burns2024weaktostrong,
  title     = {Weak-to-Strong Generalization: Eliciting Strong Capabilities
               with Weak Supervision},
  author    = {Burns, Collin and Izmailov, Pavel and Kirchner, Jan Hendrik and
               Baker, Bowen and Gao, Leo and Aschenbrenner, Leopold and
               Chen, Yining and Ecoffet, Adrien and Joglekar, Manas and
               Leike, Jan and Sutskever, Ilya and Wu, Jeffrey},
  booktitle = {Proceedings of the 41st International Conference on Machine
               Learning (ICML)},
  series    = {Proceedings of Machine Learning Research},
  volume    = {235},
  pages     = {4971--5012},
  year      = {2024},
  publisher = {PMLR}
}

@article{bowman2022scalable,
  title   = {Measuring Progress on Scalable Oversight for Large Language Models},
  author  = {Bowman, Samuel R. and Hyun, Jeeyoon and Perez, Ethan and
             Chen, Edwin and Pettit, Craig and Heiner, Scott and
             Luko{\v{s}}i{\=u}t{\.e}, Kamil{\.e} and Askell, Amanda and
             Jones, Andy and Chen, Anna and others},
  journal = {arXiv preprint arXiv:2211.03540},
  year    = {2022}
}

@inproceedings{greenblatt2024aicontrol,
  title     = {AI Control: Improving Safety Despite Intentional Subversion},
  author    = {Greenblatt, Ryan and Shlegeris, Buck and Sachan, Kshitij and
               Roger, Fabien},
  booktitle = {Proceedings of the 41st International Conference on Machine
               Learning (ICML)},
  series    = {Proceedings of Machine Learning Research},
  volume    = {235},
  pages     = {16295--16336},
  year      = {2024},
  publisher = {PMLR}
}

\appendix

\section{Complexity Analysis}
\label{sec:complexity}

We bound the cost of \ours{} by the size of the evidence graph it induces and
contrast it with the trajectory-dependent cost of a generic deep research agent.
Let $L = |R_q|$ be the number of sub-queries, $m$ the seeds retained per
sub-query, $d_{\max}$ the maximum seed degree (references plus citations in
$\mathcal{C}$), $n_{\max}$ the cap on atomic claims per paper ($n_{\max}=5$), and
$\ell$ the maximum token length of $t(p)$. We write $c_{\mathrm{plan}}$,
$c_{\mathrm{claim}}$, $c_{\mathrm{ent}}$ for the per-call cost of the planning,
claim-extraction, and entailment models, each a constant in $\ell$ and
$n_{\max}$.

\noindent \textbf{Graph size: }
With $|S| \le mL$, the node set is $|V| \le mL(d_{\max}+1) = O(mL\,d_{\max})$, and
the induced edge set is $|E| = O(|V|\,\bar{d})$ for sparse
citation graphs of average degree $\bar{d} \ll |V|$. Both bounds are fixed by
$S$ \emph{before} any inference: the exploration frontier is a deterministic
function of $S$ and the corpus, not of model judgments.

\noindent \textbf{Time:}
Stage~1 costs $O(c_{\mathrm{plan}} + L + mL\,d_{\max})$ for one planning call,
$L$ index queries, and the neighbor fetches. Stage~2 makes one claim-extraction
call per node and one entailment call per edge; each edge scores a single
$n_u\times n_v$ matrix ($n_u,n_v \le n_{\max}$) in $O(n_{\max}^2)=O(1)$ and
yields $\Delta(u,v)$ independently, so
Stage~2 costs $O(|V|\,c_{\mathrm{claim}} + |E|\,c_{\mathrm{ent}})$ plus an
$O(|V|+|E|)$ pruning pass. Stage~3 solves the PPR fixed point by power iteration on the sparse $\mathbf{W}$
rather than the $O(|V|^3)$ closed form: each
iteration is $O(|V|+|E|)$ and contracts at rate $(1-\alpha)$, so
$T=O(\alpha^{-1}\log(1/\epsilon))$ iterations suffice, with $O(|V|\log k)$ for
top-$k$ extraction. End to end,
\begin{equation}
\tiny
O\Big(
\underbrace{|V|\,c_{\mathrm{claim}} + |E|\,c_{\mathrm{ent}}}_{\text{graph weighting}}
\;+\;
\underbrace{\tfrac{1}{\alpha}\log\tfrac{1}{\epsilon}\,(|V| + |E|)}_{\text{ranking}}
\;+\; |V|\log k
\Big),
\label{eq:complexity}
\end{equation}
dominated by the $|E|$ entailment calls. The total number of model
invocations is exactly $1 + |V| + |E|$, known before any inference and
independent of intermediate outputs.

\noindent \textbf{Space: }
\ours{} stores the textual pool ($O(|V|\,\ell)$), claim sets
($O(|V|\,n_{\max})$), and weighted graph ($O(|V|+|E|)$). Each entailment matrix
is $O(1)$ and discarded once $\Delta(u,v)$ is computed, so edge scoring streams
with $O(1)$ memory per edge; power iteration adds $O(|V|+|E|)$. Total space is
$O(|V|\,\ell + |E|)$. No component maintains a growing context: peak prompt
length is $O(\ell)$ (extraction) or $O(n_{\max}^2)$ (entailment), regardless of
how many papers are processed.

\noindent \textbf{Comparison: }
A search-and-synthesis agent making $T_{\mathrm{tool}}$ tool calls, appending
each $O(\ell)$ observation to its context, incurs prompts of length $O(t\,\ell)$
at step $t$ and cumulative cost
$O(\sum_{t} t\,\ell)=O(T_{\mathrm{tool}}^2\,\ell)$, quadratic in a trajectory
length that is unbounded a priori, since the model decides termination. \ours{}
is instead \emph{linear} in a structurally fixed graph, with constant-size
prompts and a termination condition (exhaustion of $E$, then PPR convergence)
set before inference, consistent with the tool-call and token reductions in the results section.

\section{Research Plan and Seed Construction}
A complete execution trace of this stage for a real query , the
generated plan, the per-sub-query retrieval lists, and the resulting
seed set , is reproduced in Appendix~\ref{app:trace}. For that query,
\ours{} issues exactly $L=5$ external search calls, after which
no further search occurs; sub-queries whose keyword phrasing matches
papers outside the topic (e.g., quantum-measurement papers retrieved
for the phrase \emph{``faithful \ldots measurement''}) contribute seeds
that are subsequently removed by the claim-level pruning of Stage~2.

\section{Execution Trace of Stage 1}
\label{app:trace}

Trace~1 reproduces \ours{}'s Stage~1 log for one query from the
expert-curated set, lightly reformatted for readability. Orange
annotations ($\triangleleft$) connect each field to the formalism of
Section~3. The trace illustrates two properties of the pipeline. First,
the external-search budget is fixed and small: one planning call and
$L=5$ index calls, after which every subsequent operation is graph
traversal. Second, seed construction is deliberately permissive:
sub-query $\rho_1$ retrieves quantum-information and applied-physics
papers through pure keyword overlap, and these enter the seed set $S$.
Such seeds survive Stage~1 by design , lexical search over a broad
corpus cannot avoid them , and are removed in Stage~2, where every
incident edge receives coverage affinity $\Delta = 0$ and is pruned.
Relevance enforcement is thus the responsibility of the evidence graph,
not the search engine.

\begin{tracebox}{Stage 1: Plan and seed construction}
\footnotesize\raggedright\setlength{\parskip}{1.4mm}

\tkey{query:} \tval{"How to faithfully and explicitly measure the
helpfulness of human explanations to language models during finetuning
and inference?"} \tnote{input $q$}

\tkey{plan:} \tval{["faithful explicit helpfulness measurement",\,
"human explanations language models",\, "helpfulness metrics
finetuning",\, "explicit human guidance LMs",\, "measuring explanation
impact inference"]} \tnote{$R_q = \{\rho_1,\ldots,\rho_5\}$, $L=5$; one
LLM call}

\tkey{external\_calls:} \tval{Semantic Scholar API $\times$ 5}
\tnote{one per $\rho_\ell$; the only search calls in the run}

\medskip
\tkey{$\rho_2$:} \tval{"human explanations language models"}\\
\hspace*{3mm}\tkey{relevance:} \tval{Fostering Appropriate Reliance on
LLMs; Sparse Autoencoders Find Highly Interpretable Features in LMs;
FaithLM; Evaluating the Reliability of Self-Explanations in LLMs; Can
LLMs Explain Themselves?; \ldots}\\
\hspace*{3mm}\tkey{recency:} \tval{Sparse Autoencoders Find Highly
Interpretable Features in LMs; Can LLMs Explain Themselves?; Post Hoc
Explanations of LMs Can Improve LMs; \ldots}\\
\hspace*{3mm}\tkey{intersection:} \tval{Sparse Autoencoders Find Highly
Interpretable Features in LMs; Can LLMs Explain Themselves?; Exploring
LLMs for Human Mobility Prediction; Post Hoc Explanations of LMs Can
Improve LMs; Are Human Explanations Always Helpful?; \ldots}
\tnote{Seeds$(\rho_2)$, Eq.~3 , on-topic core of $S$}

\medskip
\tkey{$\rho_1$:} \tval{"faithful explicit helpfulness measurement"}\\
\hspace*{3mm}\tkey{relevance:} \tval{Exact minimum measurement
dependence for faithful local deterministic models\ldots; A resource
theory of quantum memories\ldots; FlowSteer; Pretty good measurement
for bosonic Gaussian ensembles; Topology-faithful nonparametric
estimation\ldots}\\
\hspace*{3mm}\tkey{recency:} \tval{A resource theory of quantum
memories\ldots; Improvements on sampling of point spread
function\ldots; Pretty good measurement for bosonic Gaussian ensembles;
Topology-faithful nonparametric estimation\ldots; Estimation of
non-constant variance in isothermal titration calorimetry\ldots}\\
\hspace*{3mm}\tkey{intersection:} \tval{A resource theory of quantum
memories\ldots; Pretty good measurement for bosonic Gaussian ensembles;
Topology-faithful nonparametric estimation\ldots}
\tnote{lexical-match noise: enters $S$, pruned in Stage~2 ($\Delta=0$)}

\medskip
\tkey{$\rho_3$:} \tval{"helpfulness metrics finetuning"}\\
\hspace*{3mm}\tkey{intersection:} \tval{RAFT: Reward rAnked FineTuning;
Mertech; What Happens During Finetuning of Vision Transformers; RAIN:
Your LMs Can Align Themselves without Finetuning; Cross-Domain Image
Captioning with Discriminative Finetuning; \ldots}
\tnote{mixed: alignment papers on-topic; audio/vision finetuning
tangential}

\medskip
\tkey{$\rho_4$:} \tval{"explicit human guidance LMs"}\\
\hspace*{3mm}\tkey{intersection:} \tval{Guiding Instruction-based Image
Editing via Multimodal LLMs; Recent advances in leveraging human
guidance for sequential decision-making; John praised Mary because
\_he\_?; Explicit Syntactic Guidance for Neural Text Generation;
Fine-Grained Human Feedback Gives Better Rewards for LM Training;
\ldots}
\tnote{recovers the human-feedback literature missed by $\rho_2$}

\medskip
\tkey{$\rho_5$:} \tval{"measuring explanation impact inference"}\\
\hspace*{3mm}\tkey{intersection:} \tval{Measuring the Impact of
Explanation Bias; Field-level simulation-based inference with galaxy
catalogs\ldots}
\tnote{$|$Seeds$(\rho_5)| = 2$: narrow phrasing, sparse overlap of the
two rankings}

\medskip
\tkey{seed\_set:} \tval{$|S| = 29$ papers} \tnote{union over
$\rho_1$--$\rho_5$ after deduplication; $|S| < mL$, Eq.~3}
\end{tracebox}

\paragraph{Full seed set.}
Table~\ref{tab:full-seeds} lists all 29 seed papers with their
contributing sub-query.

\begin{promptbox}{Claim Entailment Prompt}
\small

\textbf{Instruction.}
Consider the following two sentences.

\medskip
Based only on the information available in these two sentences,
which of the following options is true?

\medskip
\noindent
\textbf{a.} Sentence 1 generalizes, specifies, or has an equivalent
meaning with Sentence 2.

\smallskip
\noindent
\textbf{b.} Sentence 1 presents the reason, cause, or condition for
the result or conclusion made in Sentence 2.

\smallskip
\noindent
\textbf{c.} Sentence 2 mentions a comparison, criticism, juxtaposition,
or a limitation of something said in Sentence 1.

\smallskip
\noindent
\textbf{d.} Sentence 1 and Sentence 2 are independent.

\medskip
\noindent
\textbf{Output:} Answer with only the option letter:
\texttt{a}, \texttt{b}, \texttt{c}, or \texttt{d}.

\medskip
\noindent
\textbf{Sentence 1:} \texttt{\{premise\}}

\smallskip
\noindent
\textbf{Sentence 2:} \texttt{\{hypothesis\}}

\end{promptbox}

\begin{promptbox}{Sub-query Generation Prompt}
\small

\textbf{System Message}

\smallskip
You are an expert scientific literature retrieval assistant.
Your task is to decompose a research topic into highly relevant search
queries for academic search engines such as Semantic Scholar, \textbf{arXiv},
Crossref, and OpenAlex.

\smallskip
Your goal is to maximize retrieval of relevant papers while minimizing
irrelevant ones.

\smallskip
\textbf{Rules:}

\noindent
-- Preserve the key technical concepts from the original query.\\
-- Never invent new concepts, tasks, datasets, methods, or application domains.\\
-- Never broaden the scope of the query.\\
-- Never remove constraints that make the query specific.\\
-- Each subquery should emphasize a different aspect of the original topic.\\
-- Prefer concise keyword-style search queries rather than natural language.\\
-- Output ONLY a Python list of exactly 5 strings.

\medskip
\hrule
\medskip

\textbf{User Message}

\smallskip
Generate exactly 5 complementary search queries for the following
research topic.

\smallskip
\textbf{Requirements:}

\noindent
-- Each query should contain 2--3 technical words.\\
-- Preserve all important constraints from the original query.\\
-- Focus on different aspects of the same topic.\\
-- Do NOT introduce unrelated terminology.\\
-- Do NOT generalize the query.\\
-- Do NOT specialize beyond what is stated.\\
-- Use terminology commonly appearing in paper titles and abstracts.\\
-- Return ONLY a Python list.

\medskip
\textbf{Examples}

\smallskip
\textbf{Example 1}

\noindent
\textbf{Query:} Retrieval-augmented generation methods for medical
question answering.

\smallskip
\noindent
\textbf{Good:}\\
-- RAG medical QA\\
-- retrieval augmented clinical answering\\
-- medical knowledge retrieval LLMs\\
-- RAG healthcare QA

\smallskip
\noindent
\textbf{Bad:}\\
-- medical question answering
\textit{(drops the retrieval-augmented constraint)}\\
-- retrieval augmented generation
\textit{(drops the medical grounding)}\\
-- clinical LLM fine-tuning
\textit{(introduces fine-tuning, which was never mentioned)}

\medskip
\textbf{Example 2}

\noindent
\textbf{Query:} Reducing hallucinations in multilingual neural machine
translation.

\smallskip
\noindent
\textbf{Good:}\\
-- hallucination multilingual translation\\
-- faithfulness multilingual MT\\
-- reducing translation hallucinations\\
-- multilingual MT hallucination mitigation

\smallskip
\noindent
\textbf{Bad:}\\
-- machine translation quality
\textit{(too broad)}\\
-- LLM hallucination detection
\textit{(drops translation)}\\
-- multilingual language models
\textit{(drops both hallucination and translation)}

\medskip
\textbf{Example 3}

\noindent
\textbf{Query:} Parameter-efficient fine-tuning of vision transformers
for edge devices.

\smallskip
\noindent
\textbf{Good:}\\
-- PEFT vision transformers\\
-- ViT edge deployment\\
-- efficient ViT fine-tuning\\
-- adapter tuning ViT

\smallskip
\noindent
\textbf{Bad:}\\
-- vision transformer architectures
\textit{(drops efficiency and edge)}\\
-- edge device inference
\textit{(drops ViT)}\\
-- fine-tuning language models
\textit{(wrong modality)}

\medskip
\textbf{Example 4}

\noindent
\textbf{Query:} Research works on merging instruction-tuned and
domain-specialized LLMs.

\smallskip
\noindent
\textbf{Good:}\\
-- merging instruction-tuned LLMs\\
-- merging domain-specialized LLMs\\
-- model merging LLM specialization\\
-- instruction domain LLM merging

\smallskip
\noindent
\textbf{Bad:}\\
-- domain-specialized LLMs integration
\textit{(too vague)}\\
-- instruction-tuned domain-specialized models
\textit{(describes one model instead of merging two)}\\
-- LLM combination methods
\textit{(too generic)}

\medskip
\noindent
\textbf{Now generate exactly 5 search queries.}

\smallskip
\noindent
\textbf{Query:} \texttt{\{agentic\_query\}}

\end{promptbox}

\section{Claim Extraction}
\label{app:claim-extraction}

\ours{} represents each paper by a small set of atomic, contribution-level
claims extracted from its abstract and introduction. The prompt below constrains
the extractor to return 4--5 standalone claims, each tagged by type and anchored
to a verbatim source phrase, so that the resulting claims can be compared
across papers when building the entailment matrix of Stage~2.

\begin{promptbox}{Claim Extraction Prompt}
\small

\textbf{System Message}

\smallskip
You are a scientific claim extractor specialized in analyzing academic papers.
Your goal is to extract core claims that will later be used to build an
entailment matrix --- meaning each claim must be precise enough to be compared
against claims from other papers for evidential grounding relationships.

\medskip
\hrule
\medskip

\textbf{User Message}

\smallskip
Extract 4--5 core claims from the following paper excerpt
(abstract + introduction).

\smallskip
\textbf{Each claim must:}

\noindent
-- Be a \textbf{standalone, self-contained assertion}
(readable without the paper's context).\\
-- Be \textbf{atomic} (one idea per claim, no compound assertions).\\
-- Be tagged with one of these types:

\smallskip
\hspace*{3mm}$*$ \cat{EMPIRICAL} --- a result, measurement, or observed
phenomenon.\\
\hspace*{3mm}$*$ \cat{THEORETICAL} --- a mechanistic explanation or formal
relationship.\\
\hspace*{3mm}$*$ \cat{METHODOLOGICAL} --- a design choice with its
justification.\\
\hspace*{3mm}$*$ \cat{CONTEXTUAL} --- a gap, limitation, or motivation the
paper addresses.

\smallskip
\noindent
-- Represent claims the paper makes \textbf{as its own contribution}, not claims
attributed to prior work.\\
-- Be phrased as a \textbf{declarative sentence} (not a question or fragment).

\medskip
\noindent
Return your response as a JSON object in exactly this format:

\smallskip
{\ttfamily\footnotesize
\{\\
\hspace*{2mm}"claims": [\\
\hspace*{4mm}\{\\
\hspace*{6mm}"id": "C1",\\
\hspace*{6mm}"type": "EMPIRICAL | THEORETICAL | METHODOLOGICAL | CONTEXTUAL",\\
\hspace*{6mm}"claim": <standalone declarative sentence>,\\
\hspace*{6mm}"source\_phrase": <short verbatim phrase from the text>\\
\hspace*{4mm}\},\\
\hspace*{4mm}...\\
\hspace*{2mm}]\\
\}
}

\medskip
\noindent
Do not include any explanation or text outside the JSON object.

\medskip
\noindent
\textbf{Paper excerpt:} \texttt{\{text\}}

\end{promptbox}

\section{Test Data Preparation}
\label{app:test-data}

For the arXiv held-out set, we convert each paper into a single search-style
query using the prompt below. The prompt constrains the model to produce exactly
one broad, self-contained research query that captures the paper's primary
contribution, while varying phrasing across papers and avoiding any reference to
the title, authors, or implementation details that would leak the ground truth.

\begin{promptbox}{Query Generation Prompt (Test Data)}
\small

\textbf{System Message}

\smallskip
You are given the text of a research paper. Your task is to generate exactly one
general research query that is directly answered by the paper.

\smallskip
\textbf{Guidelines:}

\noindent
-- Generate exactly ONE query.\\
-- The query must express a SINGLE information need. Do NOT combine multiple
questions, requests, or comparisons into one query.\\
-- The query should capture the paper's primary research problem or main
contribution rather than secondary details.\\
-- The query should be broad enough that someone could naturally use it to
search for relevant research papers.\\
-- Keep the query concise, ideally between 8 and 18 words, and never exceed
25 words.\\
-- Avoid yes/no questions. Vary the phrasing across papers. Do NOT repeatedly
start queries with ``Can'', ``Could'', ``Is'', or ``Are''. Instead, naturally
vary the phrasing, for example:

\smallskip
\hspace*{3mm}$*$ How does \ldots\\
\hspace*{3mm}$*$ What are \ldots\\
\hspace*{3mm}$*$ Which methods \ldots\\
\hspace*{3mm}$*$ Why do \ldots\\
\hspace*{3mm}$*$ When does \ldots\\
\hspace*{3mm}$*$ What approaches \ldots

\smallskip
\noindent
-- Do NOT mention the paper title, authors, or implementation details.\\
-- Mention datasets, benchmark names, model names, or algorithms only if they
are central to the research question.\\
-- The query should sound like a realistic query that a researcher would type
into Google Scholar, Semantic Scholar, or an academic search engine.\\
-- Return ONLY the query. Do not prepend ``Query:'', use quotation marks, or
provide any explanation.

\medskip
\noindent
\textbf{Example}

\smallskip
\noindent
\textbf{Query:} Could you recommend a dataset paper which presents relation
extraction performance on translated data?

\smallskip
\noindent
\textbf{Research\_paper:} \texttt{\{reference\_text\}}

\smallskip
\noindent
The above research paper directly answers the example query.

\medskip
\noindent
Now generate exactly one concise, diverse, search-oriented research query for the
new research paper.

\medskip
\hrule
\medskip

\textbf{User Message}

\smallskip
\noindent
\texttt{\{text\}}

\end{promptbox}

\section{Entailment Model Details}
\label{app:entailment}

We fine-tune \texttt{Qwen2.5-3B-Instruct} on the union of MSciNLI and SciNLI,
formulated as a four-way claim-relation classification task (entailment,
reasoning, contrasting, neutral). Training uses full-parameter SFT with
DeepSpeed ZeRO-3. Table~\ref{tab:entail-hparams} lists the full configuration.

\section{Corpus Statistics}
\label{app:corpus-stats}

Table~\ref{tab:corpus-stats} reports the number of papers per arXiv category in
our corpus $\mathcal{C}$. Papers may carry multiple category tags, so the
per-category counts sum to more than the corpus size.


\begin{table}[h]
\centering
\small
\setlength{\tabcolsep}{6pt}
\renewcommand{\arraystretch}{1.1}
\begin{tabular}{lr@{\hskip 20pt}lr}
\toprule
\toprule
\textbf{Category} & \textbf{\#} & \textbf{Category} & \textbf{\#} \\
\midrule
cs.LG   & 135{,}116 & cs.NE   & 6{,}972 \\
cs.CV   & 150{,}055 & cs.MA   & 3{,}890 \\
cs.AI   & 36{,}148 & stat.ML & 18{,}549\\
cs.CL   & 82{,}273 &         &    \\
cs.IR   & 14{,}081  &         &    \\
\bottomrule
\bottomrule
\end{tabular}
\caption{Per-category paper counts in $\mathcal{C}$. Papers with multiple tags
are counted once per category.}
\label{tab:corpus-stats}
\end{table}

\begin{table}[h]
\centering
\small
\setlength{\tabcolsep}{6pt}
\renewcommand{\arraystretch}{1.15}
\begin{tabular}{ll}
\toprule
\toprule
\textbf{Setting} & \textbf{Value} \\
\midrule
Base model & \texttt{Qwen2.5-3B-Instruct} \\
Training stage & Supervised fine-tuning (SFT) \\
Fine-tuning type & Full parameters \\
Optimization backend & DeepSpeed ZeRO-3 \\
Prompt template & \texttt{qwen} \\
\midrule
Max sequence length & 2048 \\
Epochs & 4 \\
Per-device batch size & 16 \\
Gradient accumulation & 1 \\
Learning rate & $1{\times}10^{-5}$ \\
LR scheduler & Cosine \\
Warmup ratio & 0.1 \\
Precision & bf16 \\
\midrule
Validation split & 10\% \\
Eval strategy & Every 500 steps \\
Eval batch size & 1 \\
\bottomrule
\bottomrule
\end{tabular}
\caption{Fine-tuning configuration for the \ours{} entailment model.}
\label{tab:entail-hparams}
\end{table}
\section{Effect of Exploration Hop} \label{app:hop_analysis} The 1.5-hop construction expands only to immediate neighbors of the seeds but additionally retains citation relations among the retrieved nodes. We compare this design with a one-hop graph containing only seed--neighbor relations and a full two-hop expansion. \begin{table}[t] \centering \small \setlength{\tabcolsep}{5pt} \renewcommand{\arraystretch}{1.12} \begin{tabular}{lrrr} \toprule \toprule \textbf{Graph} & \textbf{R@50} & $\mathbf{|V|}$ & \textbf{Entailment calls} \\ \midrule 1-hop & 0.331 & 642 & 1,286 \\ 1.5-hop & \textbf{0.372} & 642 & 3,947 \\ 2-hop & 0.386 & 4,781 & 28,614 \\ \bottomrule \bottomrule \end{tabular} \caption{Effect of exploration radius on ICLR. The 1-hop and 1.5-hop constructions contain the same nodes; 1.5-hop additionally retains relations among retrieved neighbors.} \label{tab:hop_ablation} \end{table} The one-hop graph is cheaper but discards relations among neighboring papers, reducing Recall@50. A full two-hop expansion provides only a small additional recall gain while increasing the candidate graph and entailment workload by an order of magnitude. The 1.5-hop construction therefore captures much of the benefit of additional graph connectivity without allowing the candidate space to grow recursively.

\section{Effect of the Seed Budget} \label{app:seed_budget} The seed budget controls the trade-off between the size of the structurally bounded search space and candidate coverage. We vary the maximum number of seeds retained per sub-query while keeping $L=5$ fixed. \begin{table}[t] \centering \small \begin{tabular}{cccc} \toprule \toprule $\boldsymbol{m}$ & \textbf{R@50} & \textbf{Median $|V|$} & \textbf{Relative cost} \\ \midrule 5 & 0.326 & 371 & $0.61\times$ \\ 10 & \textbf{0.372} & 655 & $1.00\times$ \\ 15 & 0.381 & 911 & $1.42\times$ \\ 20 & 0.386 & 1,174 & $1.83\times$ \\ \bottomrule \bottomrule \end{tabular} \caption{Effect of the seed budget on ICLR.} \label{tab:seed_budget} \end{table} Increasing the seed budget improves coverage, but the gain diminishes beyond $m=10$ while graph construction and entailment cost continue to grow. We therefore use $m=10$ as a practical trade-off between retrieval quality and the size of the bounded evidence graph.

\section{Qualitative Examples of Claim Extraction}
\label{app:claim_examples}

Table~\ref{tab:claim-examples} shows representative outputs of the
claim-extraction stage for two papers with different contribution structures.
For each paper, the extractor returns a small set of atomic claims together
with a claim type and a source phrase from the input text. The examples
illustrate that the representation is not restricted to empirical results:
survey papers may be represented primarily by contextual and methodological
claims, whereas benchmark papers can contain empirical, methodological, and
other contribution-level claims. The source phrases provide a direct textual
anchor for inspecting each extracted claim.

\begin{table*}[h]
\centering
\small
\setlength{\tabcolsep}{4pt}
\renewcommand{\arraystretch}{1.12}
\resizebox{1.0\textwidth}{!}{%
\begin{tabular}{p{0.045\textwidth} p{0.165\textwidth} p{0.38\textwidth} p{0.39\textwidth}}
\toprule
\toprule
\textbf{ID} & \textbf{Type} & \textbf{Extracted claim} & \textbf{Source phrase} \\
\midrule
\multicolumn{4}{l}{\textbf{Paper 1: A Review of Cooperation in Multi-agent Learning}}\\
\midrule

C1 & CONTEXTUAL &
Cooperation in multi-agent learning intersects with game theory, economics, social sciences, and evolutionary biology. &
``Cooperation in multi-agent learning (MAL) is a topic at the intersection of numerous disciplines'' \\

C2 & CONTEXTUAL &
Research in multi-agent learning aims to understand how agents can coordinate effectively when goals are aligned. &
``Research in this area aims to understand both how agents can coordinate effectively when goals are aligned'' \\

C3 & CONTEXTUAL &
Research in multi-agent learning also aims to understand how agents may cooperate in settings with potential for conflict. &
``how they may cooperate in settings where gains from working together are possible but possibilities for conflict abound'' \\

C4 & METHODOLOGICAL &
The paper provides an overview of fundamental concepts, problem settings, and algorithms in multi-agent learning. &
``we provide an overview of the fundamental concepts, problem settings and algorithms of multi-agent learning'' \\

C5 & CONTEXTUAL &
The paper discusses open challenges in multi-agent learning to inspire new research avenues. &
``Finally we discuss open challenges in the field with the aim of inspiring new avenues for research'' \\

\midrule
\multicolumn{4}{l}{\textbf{Paper 2: Do MLLMs Capture How Interfaces Guide User Behavior? A Benchmark for Multimodal UI/UX Design Understanding}}\\
\midrule

C1 & CONTEXTUAL &
Recent studies on UI evaluation using MLLMs focus on surface-level features, neglecting the impact of design choices on user behavior. &
``they largely focus on surface-level features, overlooking how design choices influence user behavior at scale'' \\

C2 & EMPIRICAL &
WiserUI-Bench is a benchmark built on 300 real-world UI image pairs from industry A/B tests, with empirically validated winners that induced more user actions. &
``built on 300 real-world UI image pairs from industry A/B tests, with empirically validated winners that induced more user actions'' \\

C3 & METHODOLOGICAL &
WiserUI-Bench includes expert-curated key interpretations for each instance to support post-hoc understanding of why certain UI designs succeed. &
``we support this via expert-curated key interpretations for each instance'' \\

C4 & EMPIRICAL &
Experiments on WiserUI-Bench show that MLLMs exhibit limited understanding of the behavioral impact of UI/UX design. &
``models exhibit limited understanding of the behavioral impact of UI/UX design'' \\

C5 & THEORETICAL &
WiserUI-Bench is designed to foster research on leveraging MLLMs for visual design in user behavior contexts. &
``We believe our work will foster research on leveraging MLLMs for visual design in user behavior contexts'' \\

\bottomrule
\bottomrule
\end{tabular}
}
\caption{Outputs of the claim-extraction stage. Each paper is
represented by a small set of atomic, typed claims together with the source
phrase from which the claim was extracted. These source anchors make the
intermediate representation directly inspectable before claim-level
entailment is computed.}
\label{tab:claim-examples}
\end{table*}

\section{Error Analysis: Seed Corruption and Error Propagation}
\label{app:seed-corruption}
 
A central claim of \ours{} is that structural boundedness \emph{localizes}
errors: because exploration is confined to the citation neighborhood
induced by the seed set $\mathcal{S}$, a faulty seed can only affect the
graph region it induces, and claim-level pruning further suppresses the
spurious edges it introduces. We test this directly with a controlled
seed-corruption stress test that replaces the seeds of each query with
corrupted seeds of varying relatedness, holding the rest of the pipeline
fixed like 1.5-hop expansion, coverage-affinity pruning with threshold
$\tau$, and PPR ranking are all unchanged.
 
\begin{figure*}[h]
\centering
\includegraphics[width=0.80\textwidth]{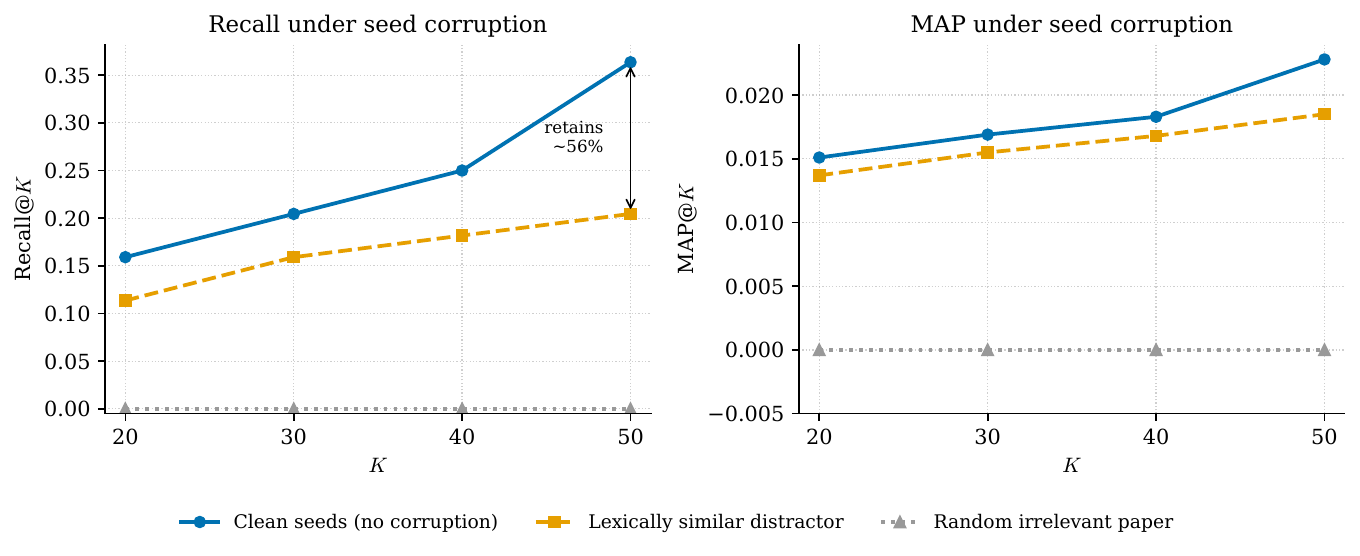}
\caption{Recall@$K$ (left) and MAP@$K$ (right) under seed corruption.
Lexically similar distractors degrade retrieval gracefully---\ours{} retains
$\sim\!56\%$ of clean Recall@50 and $\sim\!81\%$ of clean MAP@50, because
claim-level pruning suppresses the unsupported edges the distractor
introduces. Random irrelevant seeds collapse to zero: an off-manifold seed
induces a neighborhood disconnected from the ground truth, which Crase
does not attempt to recover through further search.}
\label{fig:seed-corruption}
\end{figure*}

\paragraph{Setup:}
For each query we substitute the seed papers with one of two corruption
types and re-run Stages~2--3 without modification:
\begin{itemize}
  \item \textbf{Lexically similar distractor}: a paper with high
  lexical/surface overlap with the query but absent from the ground truth.
  Such seeds lie \emph{near} the true region of the citation graph and
  mimic the lexical-match noise that keyword search returns in practice
  (cf.\ Trace~1, where phrasing such as ``faithful\,\ldots\,measurement''
  retrieves quantum-measurement papers).
  \item \textbf{Random irrelevant paper}: a paper drawn uniformly from the
  corpus, which with overwhelming probability lies in a citation region
  disconnected from the ground truth.
\end{itemize}
We compare both against the uncorrupted \textbf{clean seeds}. Results
appear Figure~\ref{fig:seed-corruption}.
 
\paragraph{Graceful degradation under near-manifold corruption:}
Replacing the seeds with lexically similar distractors reduces Recall@50
from $0.3636$ to $0.2045$: Crase still recovers $\sim\!56\%$ of the relevant
papers it retrieves under clean seeds, and preserves $\sim\!81\%$ of clean
MAP@50 ($0.0185$ vs.\ $0.0228$). Because the distractor is topically
adjacent, its 1.5-hop neighborhood still intersects the relevant region,
and any edge the distractor cannot ground receives coverage affinity
$\Delta = 0$ and is pruned. The corruption therefore lowers \emph{how many}
relevant papers enter the candidate graph but does not degrade the ranking
of those that survive, and it does not propagate beyond the injected node's
local structure. The proportional loss is largest at deep cutoffs
(Recall@50 retains $56\%$ vs.\ $71\%$ at Recall@20), indicating the
distractor mainly costs the harder-to-reach tail of the evidence set.
 
\paragraph{Bounded failure under off-manifold corruption:}
Random irrelevant seeds collapse recall to $0$ at every cutoff. This is the
\emph{predicted} cost of structural boundedness (see the cost-of-boundedness
discussion and FAQ~Q9): a seed whose citation neighborhood never reaches the
ground truth cannot be recovered, because \ours{} deliberately forgoes
open-ended re-search. Crucially, the failure is \emph{total but localizable}
-- it surfaces as an empty relevant-neighborhood that is directly
inspectable before ranking, not as silent trajectory drift that an
open-ended agent would carry into every subsequent query.
 
\paragraph{Containment versus recovery:}
Together the two settings delineate Crase's operating envelope. \ours{}
\emph{contains} errors that remain near the true evidence region, pruning
stops them from propagating but it does not \emph{recover} from errors
that push the search off the citation manifold entirely. This is the design
trade \ours{} makes: bounded cost and inspectable evidence in exchange for
adaptive re-exploration. The realistic failure mode of keyword seeding is
precisely the near-manifold case: search engines return lexical-match noise
mixed with on-topic seeds, and Table~1 already shows Crase pruning $10$ of
$29$ such artifacts at $\Delta = 0$ without recall loss. The all-random
substitution is therefore best read as a worst-case sanity bound rather than
a condition Crase is expected to encounter in deployment.
 
 
\section{Frequently Asked Questions about \ours{}}
\label{app:faq}

This section clarifies the main design choices of \ours{} and the
trade-offs introduced by structurally bounded exploration.

\begin{faqbluebox}
\footnotesize
\textbf{Q1. What does ``structurally bounded'' actually guarantee?}

Once the seed set $\mathcal{S}$ is constructed, the exploration space is
fixed by the citation graph. Subsequent model outputs may change extracted
claims and local edge scores, but they cannot issue new external searches or
recursively enlarge the candidate set. Thus, \ours{} does not guarantee that
every retrieved paper is correct; it guarantees that the search boundary,
evidence-selection operations, and stopping rule remain explicit and
inspectable.
\end{faqbluebox}

\begin{faqgreenbox}
\footnotesize
\textbf{Q2. Is auditability the same as correctness?}

No. The claim extractor and entailment model remain learned components and
can make mistakes. The key difference is that these mistakes appear as
explicit intermediate objects: extracted claims, claim-support predictions,
coverage affinities, and retain/prune decisions. In our human evaluation,
\ours{} agrees with the majority expert judgment on $84.0\%$ of sampled
citation edges. The goal is therefore not to eliminate model error, but to
make evidence-selection decisions independently inspectable.
\end{faqgreenbox}

\begin{faqorangebox}
\footnotesize
\textbf{Q3. Why allow noisy papers into the seed set?}

Stage~1 favors candidate coverage rather than final precision. Keyword-based
academic search inevitably retrieves lexical matches that are unrelated to
the intended research question. Our execution trace illustrates this with
papers on quantum measurement and other unrelated domains. Instead of asking
the search engine to make the final evidence decision, \ours{} allows such
papers to enter $\mathcal{S}$ and delegates relevance filtering to the
evidence graph. In the example trace, $10$ of $29$ seeds are subsequently
removed because they lack surviving claim-level support.
\end{faqorangebox}

\begin{faqbluebox}
\footnotesize
\textbf{Q4. Why use claim-level evidence instead of document similarity?}

Document similarity captures whether two papers discuss related topics, but
not whether one paper supports the contribution of another. This distinction
matters for citation graphs, where topical, methodological, historical, and
substantive citations all appear as the same binary relation. \ours{}
therefore represents each paper using a small set of contribution-level
claims and evaluates citation edges through claim-level support. A retained
edge can consequently be traced back to the specific claims responsible for
its weight.
\end{faqbluebox}

\begin{faqgreenbox}
\footnotesize
\textbf{Q5. Why use coverage affinity instead of the strongest matching claim?}

A single highly related claim should not make an entire citation relation
strong. Using only
$\max_{ij}M^{(u,v)}_{ij}$ would allow one matching claim pair to dominate
the edge even when the remaining contributions are unsupported. Coverage
affinity instead measures the fraction of the citing paper's claims that
receive support:
\[
\Delta(u,v)=\frac{1}{n_u}\sum_{i=1}^{n_u}s_i(u,v).
\]
This favors references that ground several principal claims while still
allowing different cited-paper claims to support different citing-paper
claims.
\end{faqgreenbox}

\begin{faqorangebox}
\footnotesize
\textbf{Q6. Can an incorrect model judgment still propagate through the graph?}

Yes, but its influence is structurally contained. A wrong claim extraction
or entailment decision can alter an edge weight and therefore affect the
final ranking. However, it cannot trigger another external search or change
the exploration frontier after seed construction. In an open-ended research
agent, an early retrieval error may modify the context, alter subsequent
queries, and redirect the remaining trajectory. In \ours{}, the error remains
localizable to a claim, edge, or ranking path within a fixed graph.
\end{faqorangebox}

\begin{faqbluebox}
\footnotesize
\textbf{Q7. Why use a 1.5-hop neighborhood rather than one or two hops?}

The 1.5-hop construction separates \emph{node expansion} from
\emph{edge completion}. \ours{} retrieves only the immediate references and
citing papers of each seed, as in a one-hop expansion, but also retains
citation relations among all retrieved nodes. It therefore exposes richer
local structure without recursively introducing another layer of papers.
This provides more evidence paths than a strict one-hop graph while keeping
the candidate boundary substantially smaller than a full two-hop expansion.
\end{faqbluebox}

\begin{faqgreenbox}
\footnotesize
\textbf{Q8. Why combine coverage affinity with recency?}

The two terms play different roles. Coverage affinity measures semantic
evidence, whereas the age term corrects a structural bias of citation
networks: older papers have had more time to accumulate citations and may
therefore receive disproportionately high random-walk mass. In
\[
w(u\!\rightarrow\!v)
=
\Delta(u,v)\operatorname{age}(v)^{-\beta},
\]
recency cannot rescue an unsupported edge because its weight remains
controlled by $\Delta(u,v)$. Conversely, an older paper with strong evidence
is down-weighted rather than discarded. Recency therefore acts as a soft
bias correction rather than a substitute for relevance.
\end{faqgreenbox}

\begin{faqorangebox}
\footnotesize
\textbf{Q9. What is the main cost of structural boundedness?}

A relevant paper outside the citation neighborhood induced by the initial
seeds cannot be recovered through later query reformulation. An open-ended
agent may, in principle, discover such a paper by continuing to search.
\ours{} deliberately exchanges this adaptive exploration for a process whose
candidate boundary and termination rule are explicit. The candidate-space
analysis in section~\ref{sec:complexity}
separates failures caused by missing candidates from those caused by evidence pruning or ranking.
\end{faqorangebox}

\begin{faqbluebox}
\footnotesize
\textbf{Q10. Why is the advantage of \ours{} smaller on the recent arXiv set?}

The main experiments show a different pattern on arXiv: the gap between
\ours{} and the strongest graph baseline is smaller than on ACL and ICLR,
and Spector2-Deepwalk is stronger at Recall@5 while \ours{} becomes stronger
at deeper recall cutoffs. One plausible explanation is that very recent
papers expose less mature citation structure, reducing the benefit of
citation-guided propagation. We treat this as a hypothesis rather than a
conclusion; the graph-statistics and candidate-coverage analyses provide a
more direct test of this explanation.
\end{faqbluebox}

\begin{table*}[h]
\centering
\small
\setlength{\tabcolsep}{4.5pt}
\renewcommand{\arraystretch}{1.15}
\begin{tabular}{p{0.12\textwidth}p{0.25\textwidth}p{0.28\textwidth}p{0.26\textwidth}}
\toprule
\toprule
\textbf{Stage} &
\textbf{Model contribution} &
\textbf{Structural control} &
\textbf{What can be inspected?} \\
\midrule

Planning &
Generates focused sub-queries &
Exactly $L$ external search calls &
Research plan and retrieved seeds \\

Graph construction &
Extracts atomic claims &
Candidate graph fixed after seed construction &
Nodes, citation edges, extracted claims \\

Evidence scoring &
Predicts local claim support &
Each citation edge scored independently &
Entailment matrix and $\Delta(u,v)$ \\

Pruning &
Provides support scores &
Fixed threshold determines edge fate &
Why each edge is retained or removed \\

Ranking &
No new evidence is generated &
Deterministic PPR/SALSA over the weighted graph &
Edge weights and final ranking scores \\

Stopping &
None &
Graph scoring completes and ranking converges &
Explicit stopping condition \\

\bottomrule
\bottomrule
\end{tabular}
\caption{\textbf{Where does model autonomy remain in \ours{}?}
Model judgments are restricted to local, inspectable operations, while
exploration, pruning, ranking, and termination remain structurally
constrained after seed construction.}
\label{tab:control-summary}
\end{table*}

\begin{table*}[h]
\centering
\setlength{\tabcolsep}{5pt}
\renewcommand{\arraystretch}{1.12}
\begin{tabular}{p{0.90\linewidth} c}
\toprule
\toprule
\textbf{Seed paper} & \textbf{Fate} \\
\midrule
\subq{$\rho_1$}{faithful explicit helpfulness measurement}
\pruned{A resource theory of quantum memories and their faithful verification with minimal assumptions} & \prunedmark \\
\pruned{Pretty good measurement for bosonic Gaussian ensembles} & \prunedmark \\
\pruned{Topology-faithful nonparametric estimation and tracking of bulk interface networks} & \prunedmark \\
\subq{$\rho_2$}{human explanations language models}
\kept{Sparse Autoencoders Find Highly Interpretable Features in Language Models} & \keptmark \\
\kept{Can Large Language Models Explain Themselves? A Study of LLM-Generated Self-Explanations} & \keptmark \\
\pruned{Exploring Large Language Models for Human Mobility Prediction under Public Events} & \prunedmark \\
\kept{Post Hoc Explanations of Language Models Can Improve Language Models} & \keptmark \\
\kept{Are Human Explanations Always Helpful? Towards Objective Evaluation of Human Natural Language Explanations} & \keptmark \\
\kept{Interpreting Language Models with Contrastive Explanations} & \keptmark \\
\kept{Using Large Language Models to Provide Explanatory Feedback to Human Tutors} & \keptmark \\
\kept{Explanations from Large Language Models Make Small Reasoners Better} & \keptmark \\
\kept{Extracting human interpretable structure-property relationships in chemistry using XAI and large language models} & \keptmark \\
\subq{$\rho_3$}{helpfulness metrics finetuning}
\kept{RAFT: Reward rAnked FineTuning for Generative Foundation Model Alignment} & \keptmark \\
\pruned{Mertech: Instrument Playing Technique Detection Using Self-Supervised Pretrained Model with Multi-Task Finetuning} & \prunedmark \\
\pruned{What Happens During Finetuning of Vision Transformers: An Invariance Based Investigation} & \prunedmark \\
\kept{RAIN: Your Language Models Can Align Themselves without Finetuning} & \keptmark \\
\pruned{Cross-Domain Image Captioning with Discriminative Finetuning} & \prunedmark \\
\kept{GenAug: Data Augmentation for Finetuning Text Generators} & \keptmark \\
\pruned{Carve3D: Improving Multi-view Reconstruction Consistency for Diffusion Models with RL Finetuning} & \prunedmark \\
\kept{Finetuning Pretrained Transformers into Variational Autoencoders} & \keptmark \\
\pruned{RetailKLIP: Finetuning OpenCLIP backbone using metric learning on a single GPU} & \prunedmark \\
\subq{$\rho_4$}{explicit human guidance LMs}
\pruned{Guiding Instruction-based Image Editing via Multimodal Large Language Models} & \prunedmark \\
\kept{Recent advances in leveraging human guidance for sequential decision-making tasks} & \keptmark \\
\kept{John praised Mary because \_he\_? Implicit Causality Bias and Its Interaction with Explicit Cues in LMs} & \keptmark \\
\kept{Explicit Syntactic Guidance for Neural Text Generation} & \keptmark \\
\kept{Fine-Grained Human Feedback Gives Better Rewards for Language Model Training} & \keptmark \\
\kept{Learning adaptive planning representations with natural language guidance} & \keptmark \\
\subq{$\rho_5$}{measuring explanation impact inference}
\kept{Measuring the Impact of Explanation Bias: A Study of Natural Language Justifications for Recommender Systems} & \keptmark \\
\pruned{Field-level simulation-based inference with galaxy catalogs: the impact of systematic effects} & \prunedmark \\
\midrule
\multicolumn{2}{l}{\textbf{Summary:}\quad
\kept{\textbf{19 seeds retained}}\, (\keptmark)\quad
\pruned{10 lexical-match artifacts pruned}\, (\prunedmark)\quad
-- all with zero surviving claim-level support} \\
\bottomrule
\bottomrule
\end{tabular}
\caption{\textbf{The seed set is noisy by design,and it does not
matter.} The complete seed set $S$ ($|S|=29$) for the query of
Trace~1, grouped by contributing sub-query $\rho_\ell$.
\keptmark~=~seed survives Stage~2 claim-level pruning and enters the
ranking walk; \prunedmark~=~every incident edge receives coverage
affinity $\Delta = 0$ and the seed is removed from the evidence graph.
Keyword search over a 500K-paper corpus inevitably retrieves
lexical-match artifacts (quantum measurement, galaxy catalogs, retail
image classification); in a search-and-synthesis loop these would enter
the context of every subsequent step, whereas in \ours{} the
evidence graph , not the search engine , enforces relevance.}
\label{tab:full-seeds}
\end{table*}
\end{document}